\documentclass{article}

\makeatletter
\newcommand{\sectionnotoc}[1]{%
  \begingroup
    \let\addcontentsline\@gobblethree
    \section{#1}%
  \endgroup
}
\makeatother

\usepackage{times}

\usepackage{multicol}
\usepackage[bookmarks=true]{hyperref}

\usepackage{todonotes}
\setuptodonotes{inline}

\usepackage{graphicx}
\usepackage{caption}
\usepackage{subcaption}

\usepackage{colortbl} 
\usepackage{booktabs}

\usepackage{enumitem}
\usepackage{wrapfig} 
\usepackage{amsmath}  
\usepackage{amsfonts} 
\usepackage{amssymb}  
\usepackage{dsfont} 
\usepackage{amsthm}

\usepackage[ruled,linesnumbered]{algorithm2e}
\usepackage{algpseudocode}
\theoremstyle{definition}

\usepackage{times}

\usepackage{lmodern}        
\usepackage[T1]{fontenc}    
\usepackage[utf8]{inputenc} 
\usepackage{microtype}

\usepackage{etoc}
\usepackage[final]{corl_2025} 

\title{Extracting Visual Plans from  Unlabeled Videos \\via Symbolic Guidance}

\author{
    \textbf{Wenyan Yang$^1$, Ahmet Tikna$^2$, Yi Zhao$^1$, Yuying Zhang$^1$,}\\
    \textbf{ Luigi Palopoli$^{2}$, Marco Roveri$^{2}$, Joni Pajarinen$^1$} \\\\
    $^1$Department of Electrical Engineering and Automation, Aalto University \\
    $^2$Department of Engineering and Computer Science, University of Trento
}

\begin{document}
\maketitle

\begin{abstract}
Visual planning, by offering a sequence of intermediate visual subgoals to a goal-conditioned low-level policy, achieves promising performance on long-horizon manipulation tasks. To obtain the subgoals, existing methods typically resort to video generation models but suffer from model hallucination and computational cost. We present Vis2Plan, an efficient, explainable and white-box visual planning framework powered by symbolic guidance. From raw, unlabeled play data, Vis2Plan harnesses vision foundation models to automatically extract a compact set of task symbols, which allows building a high-level symbolic transition graph for multi-goal, multi-stage planning. At test time, given a desired task goal, our planner conducts planning at the symbolic level and assembles a sequence of physically consistent intermediate sub-goal images grounded by the underlying symbolic representation. Our Vis2Plan outperforms strong diffusion video generation-based visual planners by delivering 53\% higher aggregate success rate in real robot settings while generating visual plans 35$\times$ faster. The results indicate that Vis2Plan is able to generate physically consistent image goals while offering fully inspectable reasoning steps.
\end{abstract}



\sectionnotoc{Introduction}

Learning versatile and long-horizon manipulation skills from uncurated and unlabeled play data remains challenging for robot learning.
The play dataset usually consists of long trajectories, which are composed of several unknown sub-skills, for solving complex and multi-stage tasks without task labels. Such task-agnostic datasets contain rich information for solving multistage and versatile tasks. However, extracting long-horizon visual plans from play data is challenging: 1) the planner needs to discover and stitch sub-skills from unlabeled data; 2) the planner should generate physically reachable subgoals for multistage tasks.

A practical approach to learn versatile skills is through semantically grounding tasks and applying hierarchical imitation learning. Recent advances in video generative models and vision-language models have shown promise as high-level planners by synthesizing intermediate visual subgoals based on language descriptions or specifying visual targets~\cite{AVDC, du2024compositional, VILP, qin2024worldsimbench, attarian2022see}. However, three core challenges limit their applicability in real-world robot learning. First, these video generative models or vision-language models conduct planning directly in pixel space or dense latent space by forecasting futures in a blackbox way~\cite{baseline_failed1, hu2024video, xing2024aid, rosete2023latent}, which struggle to reason reliably over long horizons due to model hallucination~\cite{VLM_video_halluciantions1, VLM_video_halluciantions2, VLM_video_halluciantions3, HiP, gao2024physically, sermanet2024robovqa}. Therefore, the generated physically unreachable or unrealistic visual subgoals create difficulties for the low-level controller to achieve. Second, planning via video generative models or vision-language models is extremely slow, taking seconds to generate a feasible plan, making these methods expensive to apply in real-world applications, especially real-time systems. Third, these works require heavy human efforts to label unstructured play datasets: video demonstrations must be manually labeled with natural language descriptions. Despite recent vision language models (VLMs) showing great video understanding ability, which may potentially reduce human workload, they still face challenges in reliably generating correct task descriptions~\cite{VLM_video_halluciantions1, VLM_video_halluciantions2, VLM_video_halluciantions3, VLM_video_halluciantions4}.

\begin{figure*}[thbp]
    \centering
    \includegraphics[width=0.85\textwidth]{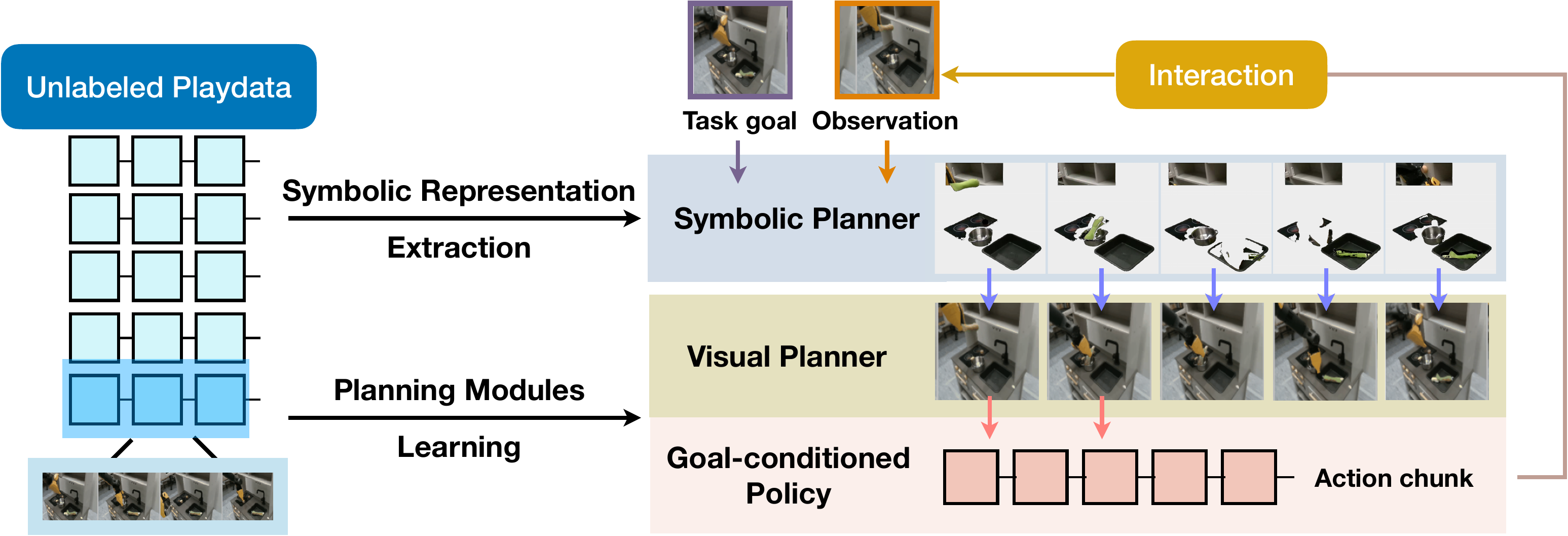}
\caption{\textbf{Vis2Plan}: Given an unlabeled play dataset, Vis2Plan  first constructs a discrete transition graph, where nodes are represented by a combination of object-centric symbols extracted from the dataset.
At inference time, given a task goal specification, a symbolic planner conducts planning at the symbolic level and generates a sequence of symbolic states, which are later converted into a sequence of visual subgoals via image retrieval and reachability estimation. These subgoals guide a goal-conditioned low-level policy to execute action chunks to achieve subgoals in a closed-loop manner.}
\label{fig:overview}
\vspace{-15pt}
\end{figure*}

To overcome these limitations, we introduce \emph{Vis2Plan}, a symbolic-guided visual planning framework that combines the representational power of vision foundation models with the efficiency and interpretability of white-box planning. Instead of training video generators, Vis2Plan first extracts a compact set of task symbols from raw, unlabeled human play data using pretrained vision foundation models (VFMs). VFMs are used as object detectors and object-centric representations to invent symbols to describe robot task states. Correspondingly, the combination of object-centric symbols serves as nodes in a discrete transition graph, capturing stable object states and the effects of primitive manipulation skills. At test time, given a desired goal symbol extracted from a goal image spedified by users, Vis2Plan then performs an A* search over the symbolic graph to find a shortest sequence of symbolic state transitions. For each symbol state in the planned sequence, we retrieve a batch of images based on each symbolic-subgoal's label from the dataset and later filter them through a reachability estimator to ensure physical achievability. The selected sequential images assemble a visual plan for the goal-conditioned low-level controller.
Finally, a goal-conditioned controller generates an h-horizon action chunk to move to the next image subgoal. The whole hierarchical planning is executed in a closed-loop manner as shown in Figure~\ref{fig:overview}.

Our approach offers several key advantages.
1) By decoupling high-level visual planning and low-level control, we reduce computational cost and eliminate the need for large, task-specific action annotations.
2) Symbolic search guarantees correctness and interpretability of the plan, and the combination of symbolic image sampling and reachability estimator prevents hallucinations and preserves photo-realism.
3) Moreover, training relies only on unlabeled play data; no human-provided task descriptions or reward annotations are required. 
We evaluate Vis2Plan in both simulation and real robotic manipulation scenarios, comparing against end-to-end goal-conditioned policies, generative video planners, and hybrid search-learn methods. In experiments with a real robot, across a suite of single- and multi-stage tasks, our framework consistently outperforms the diffusion video generative planning baseline in both subgoal completion and long-horizon success rate by 53\%. Comparing inference speed, Vis2Plan is capable of generating visual plans in an average of 0.03 seconds -- demonstrating robustness, efficiency, and fully inspectable reasoning.


\sectionnotoc{Related Work}

\textbf{Learning from unlabeled playdata} \quad
One alternative imitation learning paradigm is multi-task learning: the agent focuses on multi-task learning from play data, a form of teleoperated robot demonstration provided without a specific goal.
With play data one typically assumes that the collected demonstrations are from a rational teleoperator with possibly some latent intent in their behavior to explore the environment~\cite{playdata1}. 
Note that, unlike standard offline-RL datasets~\cite{d4rl}, play-like behavior datasets neither contain fully random actions, task descriptions, nor have rewards associated with the demonstrations~\cite{playdata1, playdata2, playdata3}. 
Previous work mostly focuses on training a end-to-end or hierarchical goal-conditioned visuomotor policy to learn the versatile behaviours~\cite{playdata1, park2023hiql}. 
Recent work proposes to leverage Vison-Language foundation models to label play data to train policy ~\cite{label_data_foundation_model} or extract a planning domains~\cite{visual-predicate, vlm_symbolics11, vlm_symbolics7, HiP}. 
However, these methods still face the trouble of long-horizon reasoning, requiring fine-tuning with specific human labelling and require high amounts of computational resources.

\textbf{Learning Abstractions from Demonstration}\quad
Learning abstractions reduce the complexity of long‐horizon planning in high‐dimensional domains~\cite{sym_plan1, sym_plan2, sym_plan3}. 
Traditional methods rely on hand‐designed abstractions~\cite{abstraction0, abstraction00}, while the PDDL format~\cite{pddl} represents actions as planning operators with preconditions and effects. 
Recent data‐driven approaches automatically discover abstractions from interactions~\cite{visual-predicator, bilevel-learn, Deepsym, abstraction2, cheng2022guided} or demonstrations~\cite{visual-predicate, im_logic1, im_logic2, im_logic3, imitation_abstraction1, imitation_abstraction2}, but assume given predicates or low‐level skill generators, limiting skill extraction from unlabeled multimodal data. 
\citet{ahmetoglu2025symbolic} learn predicates from raw data, yet only in simple environments. 
LLMs and VLMs enable high‐level planning with minimal or no demonstrations~\cite{vlm_symbolics1, vlm_symbolics2, vlm_symbolics3, vlm_symbolics4, vlm_symbolics5, vlm_symbolics6, visual-predicate}, leveraging commonsense knowledge for efficient plan generation. 
However, (1) LLM‐generated plans are challenging to refine reliably into low‐level actions~\cite{vlm_symbolics2, vlm_symbolics5}; (2) VLM‐based methods~\cite{vlm_symbolics7, vlm_symbolics1, vlm_symbolics4, vlm_symbolics9, vlm_symbolics10, vlm_symbolics11} can produce unreliable image descriptions in certain domains; and (3) their scale prohibits real‐time planning. 
In this work, we integrate symbolic planning into hierarchical visual planning by using relatively lightweight VFMs to  (i) extract symbolic transitions offline from image‐based demonstrations without prior knowledge of predicates or implemented low‐level controllers, and (ii) leverage the  symbolic skills to guide physically achievable subgoal generation instead of learning a strictly accurate neuro‐symbolic domain.

\textbf{Hierarchical  Planning from Visual Demonstrations} \quad
Hierarchical frameworks have shown promise in long‐horizon tasks. 
Recent work combines hierarchical frameworks with model‐free reinforcement learning and imitation learning to learn flexible skills from unlabeled offline data~\cite{hierarchical_latent2, latentmap, hierarchial_latent1, hierarchical_latent3, hierarchial_latent4, hierarchial_latent5, clinton2024planningtransformerlonghorizonoffline, park2023hiql, wu2024planning, li2025bilevel}. These approaches apply unsupervised learning (e.g., a VAE~\cite{hierarchical_latent2, latentmap, hierarchial_latent1, park2023hiql}) to discover abstract skill representations. 
However, these approaches require large datasets to learn a suitable skill prior and still need an additional planning model (e.g., a planner neural network), which limits the scalability and interoperability of their plans. 
Some methods combine conventional white‐box planning with learning‐based approaches~\cite{plan+learn1, plan+learn2, plan+learn3, haptic_seg} by performing search‐based planning over a learned low‐dimensional goal space and using a goal‐conditioned policy to achieve subgoals. 
Yet these methods often rely on simple goal spaces or short‐horizon tasks (e.g., Atari games~\cite{plan+learn2}; short‐horizon environments~\cite{plan+learn3}), or demand large datasets to train a goal predictor~\cite{long-visual1,long-visual2}.
From a vision‐planning perspective, several works identify temporal action abstractions based on visual similarity to detect meaningful visual skills~\cite{lotus, Bottom_up_skill}, but they cannot integrate directly with traditional search planning and still require an additional planning model for goal prediction. 
Other studies employ compositional large language foundation models and generative models to generate image sequences (videos) based on task specifications~\cite{foundation_plan1, foundation_plan2, foundation_plan3}. 
Despite impressive advances in foundation models, these approaches demand extensive language‐labeled data and computational resources, and may still fail due to generating unrealistic subgoals from limited training data. 
The key idea of Vis2Plan’s hierarchical framework is to discover symbolic temporal abstractions, enabling high‐quality, photo‐realistic, and physically reachable visual planning.

\begin{figure*}[thbp]
    \centering
    \includegraphics[width=\textwidth]{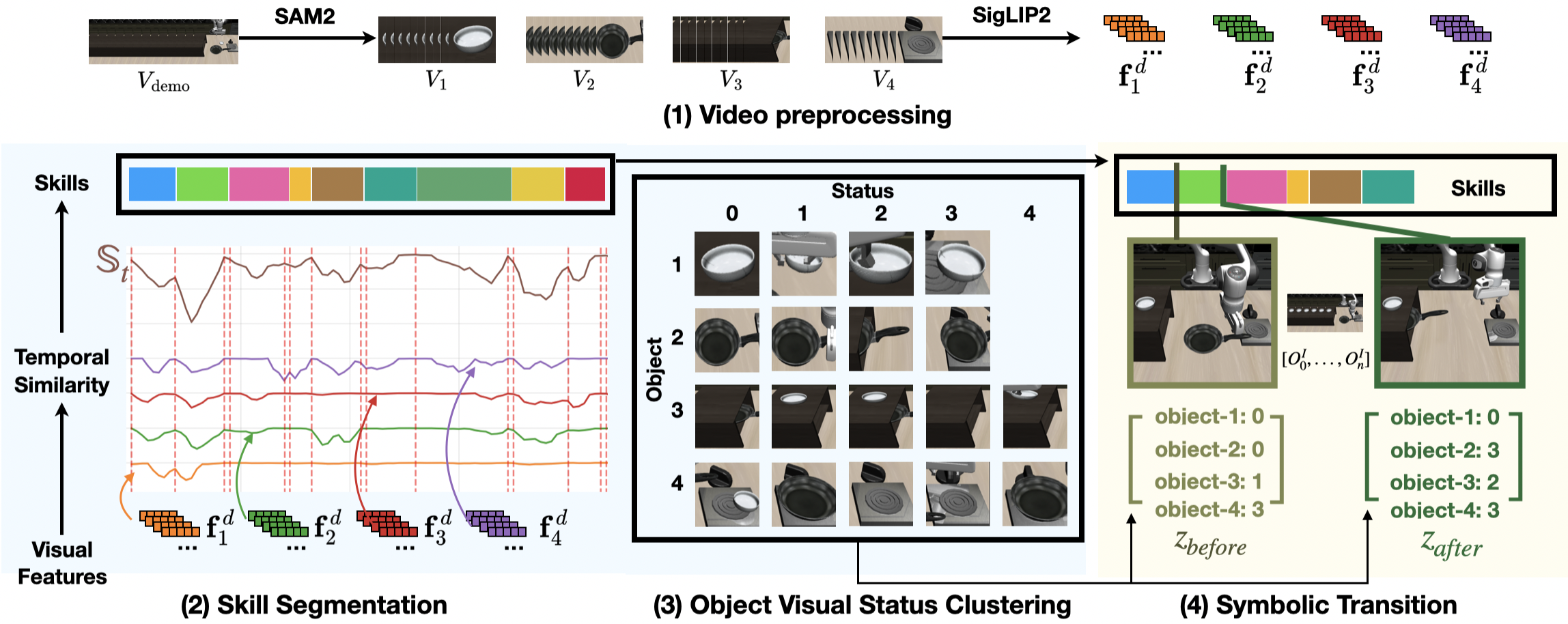}

    \caption{\textbf{Processing: symbolic skill extraction}. First, we convert the unlabelled demonstration videos into object-centric feature sequences. Then we 1) segment the demonstration into sub-skills based on object-centric temperal similarities and 2) discover the representative status of each objects. Finally, based on the clustering results and skill segmentation results, we can convert the demonstration video into symbolic transitions.}
    \label{fig:extract_skills}
    \vspace{-10pt}
\end{figure*}

\sectionnotoc{Method: Vis2Plan}

Our framework comprises three key stages: 1) Coarse symbolic skill extraction, where we leverage vision foundation models to extract symbolic skills from unlabeled video demonstrations; 2) Planning modules learning, where we utilize the extracted symbolic representations and offline dataset to train three modules (symbolic predictor, low-level controller and reachability estimator) for hierarchical visual planning; and 3) Symbolic-guided Visual Planning, the inference stage where we assemble these modules to perform closed-loop hierarchical planning and control. 

\textbf{Problem Setting: Learning from unlabeled demonstrations} \quad
We leverage human play data, where a teleoperator explores the scene (e.g., opening an oven, picking up a pot) driven by curiosity without instructions. 
Such play data provides rich state transitions and implicit knowledge of objects’ affordances and part functionalities~\cite{Mimicplay, playdata1}. 
More importantly, there is no human labeling work to identify or describe tasks. 
The unlabeled play dataset is formulated as
$\mathcal{D} = \{(o,a)\} \subset \mathcal{O} \times \mathcal{A}$
of observation–action sequences, our goal is to extract a hierarchical planner capable of handling multi-stage, multi-goal tasks. 
We also aim for the planner to discover sub-skills with symbolic abstractions, enabling it to “controllably generate” desired plans and minimize extra human annotation and curation.

\subsection{Stage 1: Coarse Symbolic Skill Extraction}
\label{sec:extract_skills}

Extracting long-horizon visual plans from unlabeled play data is challenging due to:: 1) each play sequence representing a long-horizon manipulation composed of several unknown sub-skills in random orders, and the planner needs to extract and stich subgoals from different trajectories to form a feasible plan. 2) The observation space being continuous and high-dimensional. To handle these challenges, we should a) reduce the planning horizon by discovering sub-skills and b) simplifying and discretizing the observation space. Thus we designed a two-step process to extract the symbolic representation of the task: 1) \textbf{Stable state identification}, where we aim to divide long-horizon unlabeled demonstrations into sub-skills by identifying key stable states, and 2) \textbf{Symbolic state labelling}, where we convert key image observations into discrete states described by symbols.

\textbf{Stable State Identification}\quad 
First, we preprocess demonstration videos by detecting and tracking task-relevant objects using Qwen~\cite{QWEN} and SAM2~\cite{SAM2}, creating object-centric feature sequences  (encoded with SigLIP2~\cite{siglip2}). Figure~\ref{fig:extract_skills} (1) illustrates this procedure. We then identify the stable states, where all objects’ visual status remain static or similar, and a skill is a transition between two stable states. We  compute the temporal similarity between object-centric consecutive frames using cosine similarity of object-cenrtic features, applying Gaussian smoothing, and selecting peaks through non-maximum suppression ((Figure~\ref{fig:extract_skills} (2))~\cite{motion_segment}. The peaks identify the representative stable states.  These stable states naturally divide long-horizon demonstrations into manageable sub-skills, with each sub-skill defined as a transition between two stable states. We refer to appendix for more details.

\textbf{Symbolic Transition Labelling}\quad
Next, we want to convert the identified stable states into discrete symbolic transitions. Recent studies have explored that VFMs can serve as a foundation for clustering, yielding informative groupings of embeddings~\cite{dino_classification1, dino_classification2, dino_classification3}. These findings suggest VFMs can discover representative object states in an unsupervised manner. We apply Agglomerative Clustering~\cite{agglo_cluster} with cosine similarity to extract each object's representative visual states. The number of clusters is determined via the Silhouette score~\cite{rousseeuw1987silhouettes}. Figure~\ref{fig:extract_skills} (3) illustrates examples of each object's representative states. Once the unsupervised representative states are discovered, we then train KNN classifiers to label the image frames: we take each sub-skill's precondition (start) and effect (end) frames, and use KNN classifiers to describe each frame's state. Consequently, we obtain several symbolic transitions $(z_\text{before}, z_\text{after})$ (Figure~\ref{fig:extract_skills} (4)), which can be used to construct a directed graph for in-domain planning.

\subsection{Stage 2: Planning Modules Learning}
\begin{figure*}[thbp]
   
    \centering
    \includegraphics[width=1\textwidth]{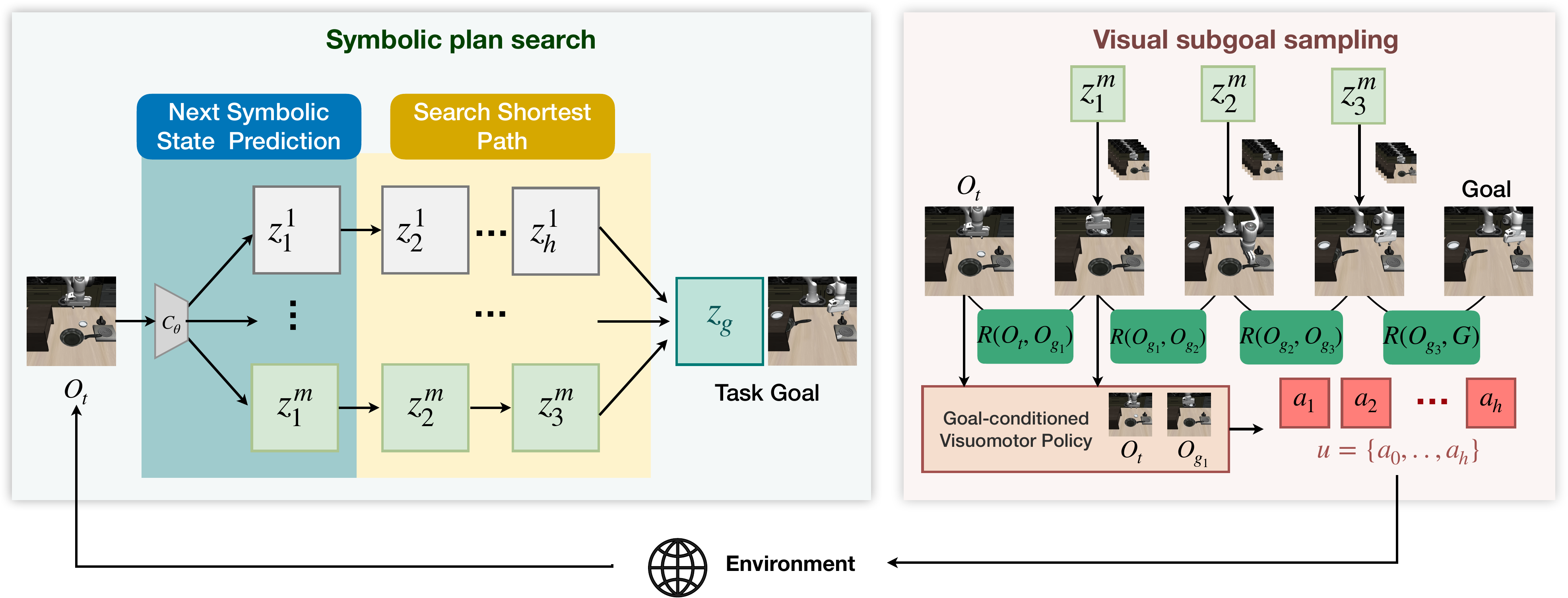}

\caption{\textbf{Symbolic‑Guided Visual Planning Framework:} Given an observation \(O_t\), the state predictor \(C(O_t)\) outputs a candidate set of next symbolic states \(\mathcal{Z} = \{z^1, \dots, z^m\}\). Using the user‑specified task goal \(z_g\), Vis2Plan searches \(\mathcal{Z}\) to find the shortest symbolic path from the current state to \(z_g\). Conditioned on this path, Vis2Plan then samples physically reachable subgoal images from the dataset to assemble a visual plan. This plan guides the goal‑conditioned controller to successfully execute the task.}

\label{fig:symbolic_guided_frame_generation}
\vspace{-10pt}
\end{figure*} 

Based on the symbolic transitions extracted from Stage 1, we can construct a symbolic graph that enables the agent to plan multistage, long-horizon tasks. However, it cannot yet support closed-loop control because: 1) the symbolic extraction stage omits intermediate frames between key states, which contain important information for closed-loop planning; 2) in practice, we found visual subgoals are critical for the low-level policy to execute the plan. Therefore, we train two modules in this stage: i) a next-symbolic-state predictor that predicts the possible next-step symbolic subgoals based on real-time visual feedback, and ii) a reachability estimator that finds sequential images matching symbolic-subgoals to construct a visual plan.

\textbf{Next-Symbolic-State Predictor} \quad
After Stage 1, the planner only has coarse symbolic transitions and cannot yet plan in image space: it must map image observations into symbolic states. Prior works~\cite{sym_plan4, sym_plan5, sym_plan6} typically train classifiers on images to  predict the action phase: "precondition" (state before action execution), "effect" (state after action execution) or "execution" (intermediate states) of each symbolic transition~\cite{sym_plan6}. Yet these methods can be inaccurate. Here, given an image observation $O_t$, we train a classifier-like predictor $C_{\theta}$ to predict the next symbolic state. During training, we sample symbolic transitions $\{z_{\mathrm{before}},z_{\mathrm{after}}\}$ and their intermediate frames $[O^I_0,\dots,O^I_n]$. We optimize $C_{\theta}$ with a cross-entropy loss to predict $z_{\mathrm{after}}$ from any intermediate $O^I_i$:
\[
\min_{\theta}\,-\mathbb{E}_{z_{\mathrm{after}},O^I_i}\bigl[\log C_{\theta}(z_{\mathrm{after}}\mid O^I_i)\bigr].
\]

\textbf{Reachability Estimator}\quad
To ensure image subgoals are physically feasible, we train a reachability estimation function $R_\psi$ to measure reachability between visual subgoals. We adopt contrastive reinforcement learning (CRL)~\cite{CRL1, CRL2, CRL3}, which leverages contrastive learning to estimate the discounted state occupancy measure. We  modify the MC-InfoNCE loss~\cite{CRL1} to the state-only case:
\begin{equation}
\mathcal{L}_{\mathrm{MC\text{-}InfoNCE}}(R_\psi)
\mathbb{E}{
\substack{
(o)\sim p(o),
o{t+}^{(1)}\sim p^\pi(o_{t+}\mid o),
o_{t+}^{(2)}\sim p(o)
}
}
\left[
\log
\frac{
\exp\bigl(R_\psi(o,o_{t+}^{(1)})\bigr)
}{
\sum_{i=1}^{N}\exp\bigl(R_\psi(o,o_{t+}^{(i)})\bigr)
}
\right],
\end{equation}
where $p(o)$ is the visual observation distribution of unlabeled play dataset
$\mathcal{D}$, and $o_{t+}^{(1)}\sim p^\pi(o_{t+}\mid o)$ denotes hindsight sampling of future observations. Since our $\pi_\psi$ imitates the offline dataset policy, this state-only on-policy MC-InfoNCE loss can estimate the state-occupancy measure of $\pi_\phi$. Consequently, $R_\psi(o_i,o_j)$ serves as a reachability score: higher values indicate better physical reachability from $o_i$ to $o_j$ for $\pi_\phi$ based on offline dataset. We also calculate a reachability threshold $\delta$ here: $ \delta=\text{min}(R_\psi(o_t, o_{t+}))$. 

For the low-level policy, we implement a goal‑conditioned visuomotor policy $\pi_{g}$~\cite{Adaflow, Mimicplay} as our low‑level controller. Given the current observation $O_t$ and the next image goal $O_{t+h}$ from a video plan, $\pi_{g}$ outputs an action chunk $a_{t:t+h}$ that advances toward $O_{t+h}$, which the agent then executes for a fixed number of timesteps.  
We train $\pi_{g}$ on paired snippets $(O_{i:i+h},a_{i:i+h})$ by sampling a random start index $i$ and horizon $h$, and optimizing:
\begin{equation}
\max_{\phi}\,\mathbb{E}_{(O_{i:i+h},a_{i:i+h})\sim D}\bigl[\log\pi_{\phi}(a_{i:i+h}\mid O_i,O_{i+h})\bigr].
\end{equation}
Both $i$ and $h$ are random: $i$ selects a timestep within an episode, and $h$ defines the policy horizon.

\subsection{Stage 3: Symbolic-guided Visual Planning}
\label{sec:stage3}
\begin{figure*}[thbp]
    \centering
    \includegraphics[width=\textwidth]{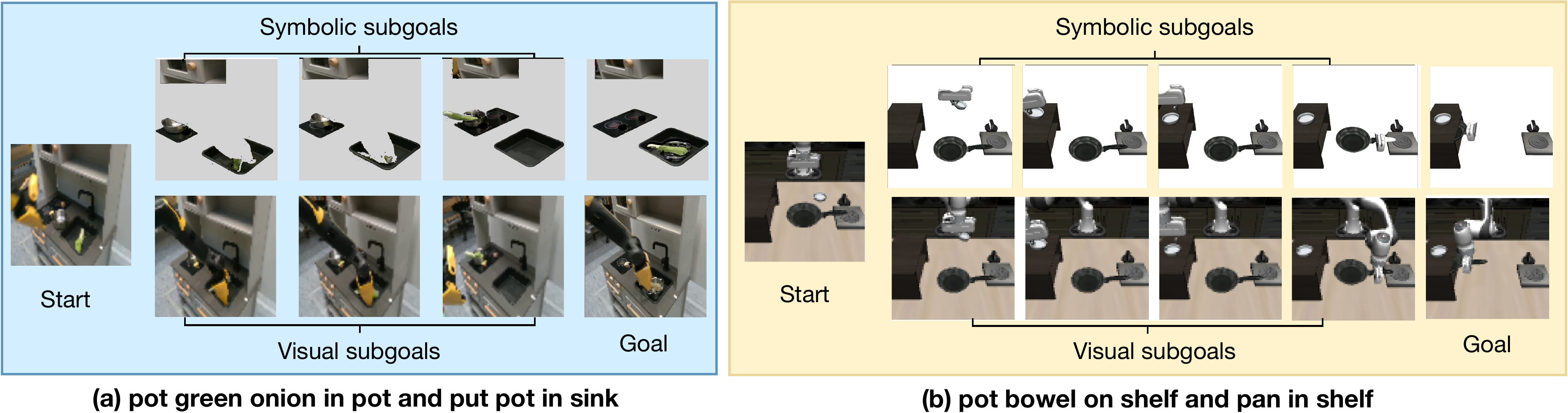}
\caption{Examples of Vis2Plan's visual plan: a real robot kitchen task (left) and a LIBERO simulation kitchen task (right).}

\label{fig:vis_plan_example}
\end{figure*}

Combining the modules learned from stage 2, we now have all the ingrediantes for Vis2Plan. To plan to achieve certain specified task goal, Vis2Plan will make three-stage planning: \textbf{1) symbolic planning: } given current image observation $O_t$ and task symbolic goal, $C_\theta$ will predict a set of next symbolic state candidates $Z_\text{next} = \{z^1_1, ..., z^m_1\}$. By applying A* search over the symbolic graph for each candidate in $Z_\text{next}$, we choose the shortest path plan as symbolic subgoal sequence $\tau_z=\{z_1, ..., z_H,z_g\}$. \textbf{2) Visual subgoal generation: }  For each symbolic subgoal $z_i$ in $\tau_z$, we can sample $z_i$ corresponding visual observation  based on the relabelled dataset \(O_i \sim P_D\bigl(\,\cdot\mid z_i\bigr)\). Given the sampled symbolic subgoal related images, the planner use beam-search to find a sequence of of visual subgoal $\tau_O=\{O_1,..,O_h\}$ that can maxmize this objective
\begin{equation}
\label{eq:hybrid_opt}
\arg\max_{\tau_O}
R(\tau_O)
\;=\;
- \sum_{i=0}^{H-1} R_\psi(O_i, O_{i+1})
\quad\text{s.t.}\quad
R_\psi(O_i, O_{i+1})\le \delta,\ \forall i,
\end{equation}
\textbf{3) Goal-conditioned action chunk generation: } Based in the $\pi_g$ will generate a h-horizon action chunk $u = \{a_0, .., a_h\}
$ based on the current visual observation $O_t$ and the next visual subgoal $O_g$: $u=\pi_g(O_t, O_g)$. Figure~\ref{fig:vis_plan_example} presents examples of visual plans.

\sectionnotoc{Experiments}

\textbf{Setting}\quad We evaluate the methods in both simulation (LIBERO Mimicplay kitchen~\cite{liu2024libero, Mimicplay}) and a real world task (kitchen manipulations). Tasks are categorized based the number of subgoals: single-stage tasks (e.g., grasp the bowel) 
and multi-stage tasks (e.g., grasp the bowel and put it on the stove). The sub-task horizon is between 140 and 400 action steps (short-horizon), and approximately between 500 ad 1400 action steps for multi-stage tasks  (long-horizon) with a 10Hz control frequency. We use the play data from~\cite{Mimicplay} as offline dataset for simulation tasks. For the real robot experiments, we collected 15 teleoperated human play trajectories, each trajectory potentially contain 5 to 15  sub-tasks. We use the task success rate as an evaluation metric.  For more details about the environments and simulation results, please refer to the Appendix.

\textbf{Baselines}  \quad
\textbf{1) End‑to‑end goal‑conditioned policies:} Goal‑conditioned (GC) policies naturally handle multiple tasks and can be trained via goal‑conditioned behavior cloning with hindsight relabeling. These serve as our low‑level controllers. We evaluate four GC‑BC variants: \textit{GC‑RNN}, \textit{GC‑Transformer},\textit{GC‑Diffusion} from ~\cite{Mimicplay} and  \textit{GC‑ACT}, a image goal-conditioned variant of \cite{aloha}.
\textbf{2) Hierarchical planners:}  each of these methods will have a high-level visual planner generating image subgoals and a low-level goal-conditioned visuomotor policy (GC-diffusion ). We implemented a) \textit{AVDC}~\cite{AVDC}, which trains a video generative diffusion model to make a visual plan.
b) graph‑search retrieval (\textit{GSR})~\cite{gsr}, which builds a graph over image latents and connect two vertex if their latent features are similar. c) \textit{UVD‑graph}, which uses UVD~\cite{UVD} to segment long‑horizon tasks into subgoals, and then build a graph similarly as GSR. For hierarchical planner baselines, We adopt Adaflow~\cite{Adaflow}, a flow‑based generative model, as low-level policy for LIBERO simulation tasks; and  we use Mimicplay policy ~\cite{Mimicplay} for real robot tasks.

Additionally, we note that methods in~\cite{im_logic1,baseline_failed1,baseline_failed2}, despite sharing similar hierarchical control methodologies, failed on all tasks and are thus omitted from further comparisons. 

\paragraph{Performance in Simulation}
\begin{table*}[h!]
\centering
\caption{Performance comparisons in Mimicplay LIBERO simulation environment. Results show success rates for both single-goal (short-horizon) and multi-goal (long-horizon) tasks with standard deviations across 5 seeds. Bold values indicate highest performance. Our Vis2Plan consistently outperforms baselines, particularly excelling in challenging long-horizon tasks where end-to-end and other hierarchical approaches struggle.}
\label{tab:performance_comparison}
\makebox[\textwidth][c]{%
  \resizebox{1.03\textwidth}{!}{%
    \begin{tabular}{l*{9}{c}}
      \toprule
      & \multicolumn{4}{c}{\textbf{Short Horizon ($1$ goal)}} 
      & \multicolumn{5}{c}{\textbf{Long Horizon ($\ge2$ subgoals)}} \\
      \cmidrule(lr){2-5} \cmidrule(lr){6-10}
      & Task-1 & Task-2 & Task-3 & Task-4 
      & Task-1 & Task-2 & Task-3 & Task-4 & Task-5 \\
      \midrule
      GC-RNN         & $\mathbf{0.96}_{\pm0.02}$ & $0._{\pm0.}$ & $0._{\pm0.}$ & $0._{\pm0.}$ & $0._{\pm0.}$ & $0._{\pm0.}$ & $0._{\pm0.}$ & $0._{\pm0.}$ & $0._{\pm0.}$ \\
      GC-Transformer & $0.95_{\pm0.03}$          & $0.01_{\pm0.02}$ & $0._{\pm0.}$ & $0.01_{\pm0.02}$ & $0._{\pm0.}$ & $0._{\pm0.}$ & $0._{\pm0.}$ & $0._{\pm0.}$ & $0._{\pm0.}$ \\
      GC-Diffusion   & $0.70_{\pm0.05}$          & $0.81_{\pm0.03}$ & $0.92_{\pm0.03}$ & $0.34_{\pm0.03}$ & $0.02_{\pm0.00}$ & $0.04_{\pm0.00}$ & $0.14_{\pm0.00}$ & $0.28_{\pm0.00}$ & $0.14_{\pm0.00}$ \\
      GC-ACT         & $0.95_{\pm0.03}$          & $0.01_{\pm0.02}$ & $0._{\pm0.}$ & $0.01_{\pm0.02}$ & $0._{\pm0.}$ & $0._{\pm0.}$ & $0._{\pm0.}$ & $0._{\pm0.}$ & $0._{\pm0.}$ \\
      \midrule
      GSR            & $0._{\pm0.}$              & $0.81_{\pm0.04}$ & $0.00_{\pm0.}$ & $0.35_{\pm0.10}$ & $0._{\pm0.}$ & $0._{\pm0.}$ & $0._{\pm0.}$ & $0._{\pm0.}$ & $0._{\pm0.}$ \\
      UVD-graph      & $0.58_{\pm0.15}$          & $0.36_{\pm0.10}$ & $0.00_{\pm0.04}$ & $\mathbf{0.36}_{\pm0.05}$ & $\mathbf{0.18}_{\pm0.03}$ & $\mathbf{0.24}_{\pm0.09}$ & $0._{\pm0.}$ & $0._{\pm0.}$ & $0._{\pm0.}$ \\
      AVDC           & $0.71_{\pm0.11}$          & $0.26_{\pm0.12}$ & $0.48_{\pm0.09}$ & $\mathbf{0.36}_{\pm0.05}$ & $0.00_{\pm0.11}$ & $0.04_{\pm0.15}$ & $0.66_{\pm0.10}$ & $0.54_{\pm0.08}$ & $0.46_{\pm0.09}$ \\
      \midrule
      Vis2Plan (ours)& $0.78_{\pm0.10}$          & $\mathbf{0.94}_{\pm0.04}$ & $\mathbf{0.94}_{\pm0.05}$ & $0.32_{\pm0.11}$ & $0.14_{\pm0.10}$ & $\mathbf{0.24}_{\pm0.09}$ & $\mathbf{0.72}_{\pm0.10}$ & $\mathbf{0.56}_{\pm0.09}$ & $\mathbf{0.82}_{\pm0.08}$ \\
      \bottomrule
    \end{tabular}%
  }
}
\end{table*}

We first compare Vis2Plan with baseline approaches in the LIBERO simulation environment~\cite{Mimicplay, liu2024libero}. Table~\ref{tab:performance_comparison} reports the task success rate of our approach Vis2Plan and baselines. First, the results indicate that end-to-end goal-conditioned approaches such as GC-RNN and GC-Transformer generally struggle, only accomplishing the short-horizon Task-1 (turning on the stove knob). However, GC-Diffusion achieves modest performance on some tasks: 0.81 on Task-2 subgoal and up to 0.28 on a long-horizon task. Based on the end-to-end baseline results, we opt to use GC-diffusion as the low-level goal-conditioned visuomotor policy for all hierarchical frameworks. Hierarchical solutions such as GSR and UVD-graph show reasonable subgoal-level results but suffer from inaccurate high-level subgoal planning, leading to a noticeable drop in success rates for longer tasks. In contrast, both mimicplay and diffusion variants of Vis2Plan consistently outperform other baselines, attaining high success rates across most subgoal tasks: 0.95 and 0.94 on Task-1 and Task-3. Vis2Plan especially excels on challenging long-horizon scenarios, yielding 0.59 and 0.82 on Task-4 and Task-5, respectively. 
Comparing the end-to-end GC-Diffusion with GC-Diffusion integrated hierarchical planners, we can see that AVDC and our Vis2Plan improve the task performance over solely using GC-Diffusion, which indicates meaningful visual subgoal plans can guide to better task success rates. In the following paragraph, we will visual plan quality analysis which discovers how high-level planners affect short/long-horizon task performances.



\begin{wraptable}{r}{0.6\linewidth}
  \vspace{-1pt}
  \centering
  \caption{\textbf{Real robot experiment.}  Success rate (over 11 trials) for each method across six manipulation tasks, grouped by short-horizon (Tasks 1–3) and long-horizon (Tasks 4–6) scenarios. Vis2Plan achieves the highest average performance.}
  \label{tab:real}
  \resizebox{\linewidth}{!}{%
    \begin{tabular}{l*5c}
      \toprule
       & \textbf{GC-Diffusion} & \textbf{GSR} & \textbf{UVD-graph} & \textbf{AVDC} & \textbf{Vis2Plan} \\
      \midrule
      \textbf{Task-1 (Short)} & 0.0 & 0.45 & 0.09 & 0.0  & {1.0}  \\
      \textbf{Task-2 (Short)} & 0.0 & 0.65 & 0.18 & 0.09 & 0.64         \\
      \textbf{Task-3 (Short)} & 0.0 & 0.55 & 0.55 & 0.55 & 0.45         \\
      \midrule
      \textbf{Task-4 (Long)}  & 0.0 & 0.45 & 0.36 & 0.18 & 0.55         \\
      \textbf{Task-5 (Long)}  & 0.0 & 0.55 & 0.55 & 0.27 & {0.63}\\
      \textbf{Task-6 (Long)}  & 0.0 & 0.09 & 0.09 & 0.0  & {1.0} \\
      \midrule
      
      \textbf{Average}  & 0.0 & 0.47 & 0.30 & 0.18  & \textbf{0.71} \\
      \bottomrule
    \end{tabular}%
  }
\vspace{-10pt}
\end{wraptable}

\paragraph{Real Robot Experiments} Table~\ref{tab:real} reports real-world performance in short-horizon (Tasks1--3) and long-horizon (Tasks4--6) scenarios. In this real setup, we select the hierarchical planning frameworks GSR, UVD-graph, AVDC, and GC-diffusion  as baselines based on their higher simulation success rates. However, we noticed that the GC-diffusion policy faces challenges in low-level control due to potentially limited training data. 
Instead, we use Mimicplay policy~\cite{Mimicplay} with sub-trajectory retrieval strategy: it retrieves a sequence of end-effector poses based on the visual subgoals, and feedd the visual-subgoals and the retrieved end-effector pose trajectory to Mimicplay policy for control. We present the real experiment results in Table~\ref{tab:real}. 
GC-Diffuser, which performed moderately well in simulation, fails to complete any task in the real world, likely due to limited demonstration data. In contrast, our Vis2Plan approach either matches or exceeds the best-performing methods in five out of six tasks, achieving perfect success (1.0) on Task1 and Task6 while outperforming all baselines on Task5. Although our method achieves slightly lower results for Task3, it remains competitive and demonstrates overall robustness and scalability for both short and long task horizons.

\begin{figure*}[thbp]
    \centering
    \includegraphics[width=1\textwidth]{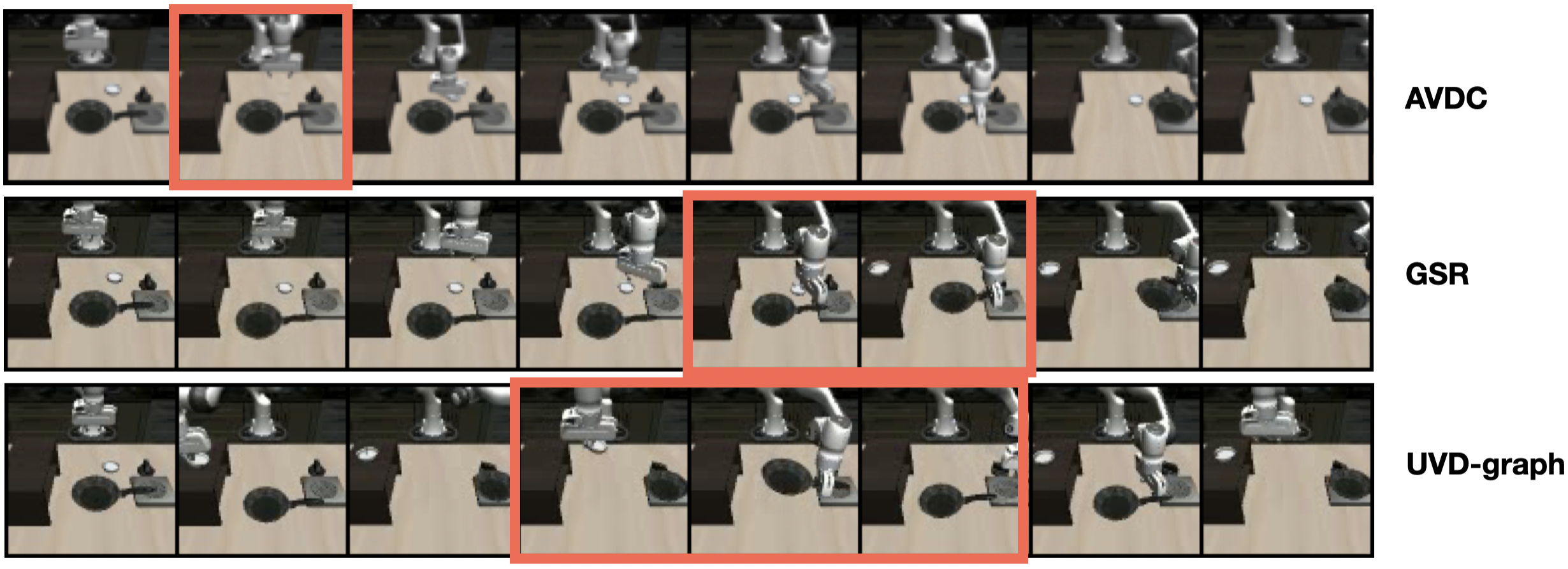}
\caption{Baseline subgoal generation failure examples. The red bounding box highlights the physically unreachable frames: the bowel disappreared in AVDC generation and UVD-graph; For GSR plan, the bowel's location changed while gripper is manipulating pan. }
\vspace{-15pt}
\label{fig:failure_example}
\end{figure*}

\textbf{Visual Plan Quality Analysis} \quad We now analyze why the baseline hierarchical planners fail. One assumption is the generated subgoal plans may not be physically achievable or photo-unrealistic. In our evaluation protocol, we allowed each planner to generate up to 20 plan sequences (represented as image frames) and manually inspected whether each sequence was ``meaningful.''. We use the reachability estimator $R_\psi$ and the reachability threshold $ \delta$ in Eq.~\ref{eq:hybrid_opt} to dertmine if the plan is meaningful: every adjacent subgoal pair should be physically achievable. The primary evaluation metric was the high-level plan success rate (the ratio of making ``meaningful.'' visual plans).
\begin{wraptable}{r}{0.5\linewidth}
  \vspace{-0pt}
  \centering
  \caption{\textbf{High-level plan quality analysis.} Success rates for single- and multi-task scenarios in simulation and real setups.}
  \label{tab:performance_comparison}
  \resizebox{\linewidth}{!}{%
    \begin{tabular}{l*{4}{c}}
      \toprule
       & \textbf{GSR} & \textbf{UVD-graph} & \textbf{AVDC} & \textbf{Ours} \\
      \midrule
      \textbf{Sim.\,1 Task}      & 0.23 & 0.25 & 0.74 & \textbf{1.00} \\
      \textbf{Sim.\,$\ge$2 Tasks}& 0.00 & 0.50 & 0.48 & \textbf{1.00} \\
      \midrule
      \textbf{Real\,1 Task}     & 1.00 & 0.75 & 0.48 & \textbf{1.00} \\
      \textbf{Real\,$\ge$2 Tasks}& 0.57 & 0.86 & 0.16 & \textbf{1.00} \\
      \bottomrule
    \end{tabular}%
  }
\end{wraptable}
Table~\ref{tab:performance_comparison} shows a comparative analysis of different methods' high-level plan quality in the real world setting, distinguishing between single-task (1Task) and multi-task ($\geq$2Tasks) scenarios. While {GSR} achieved perfect scores (1.0) for single-task plans, it often incorrectly connected visually similar states, resulting in lower performance (0.57) in multi-task settings. {AVDC} suffered from generative inconsistencies -- such as nonexistent objects including imagining two robot arms instead of the actual single arm, or missing items -- causing its score to drop significantly when tasked with multiple goals. {UVD-graph} demonstrated a reasonable overall success rate (0.75 single-task, 0.86 multi-task); however, it frequently planned overly complicated subgoals, producing prolonged image sequences where lower-level goal-conditioned policies were prone to failure. In contrast, our {Vis2Plan} approach consistently attained perfect scores (1.0) across all tasks, underscoring its robustness in producing precise, feasible, and coherent high-level plans -- a crucial requirement for reliable execution in real-world robotics applications. Figure~\ref{fig:failure_example} presents some typical failure cases observed in the baseline visual planning results.

\sectionnotoc{Conclusion}
In this paper, we introduced Vis2Plan, a symbol‑guided visual planning framework for multi‑stage robotic manipulation from unlabeled play data. Given unlabeled pay data, Vis2Plan first extracts discrete symbolic-level task abstractions and constructs a symbolic transition graph with pre‑trained vision foundation models. The symbolic transition graph enables A* search to derive high-level plans over these abstractions. Conditioning on the planned symbol trajectories, we retrieve and filter images from the dataset to construct physically feasible visual subgoals to be followed by a low-level goal-conditioned policy. We demonstrate that combining unsupervised symbolic extraction from video‑based play data with white‑box planning leads to strong performance in multi‑stage robot tasks. Our experiments -- both on the LIBERO simulation benchmark and on a real robotic manipulator -- demonstrate robust, interpretable, and efficient performance, paving the way for scalable, inspectable robot learning from unlabeled data

\sectionnotoc{Limitations}
From the high-level planning perspective, one limitation is that task-related object names in play data need to be specified by humans. Additionally, the extracted symbolic information still lacks the semantic richness of natural language descriptions, making it unable to generalize to out-of-domain settings. However, a promising research direction would be combining this framework with Vision-Language Models (VLMs) to address their complementary weaknesses: enhancing the generalization ability of Vis2Plan while mitigating the hallucination issues of VLMs. From the robot control perspective, Vis2Plan's performance bottleneck lies in the data-driven low-level goal-conditioned policy. Improving performance would require either collecting more training data or specifically designing task-appropriate low-level skill controllers.


\bibliography{reference_others}  


\clearpage

\appendix

\tableofcontents

\section{Experiment details}
\subsection{Simulation environment}

\subsubsection{Setups}
The simulation tasks are selected from the Kitchen Scene 9 environment (KITCHEN\_SCENE9) of the LIBERO benchmark~\cite{liu2024libero}, which is built on robosuite and MuJoCo. We choose LIBERO for its BDDL‐based goal specification language, which enables concise definitions of diverse tasks and facilitates unified multitask evaluation when learning from play data.

\subsubsection{Tasks}

\paragraph{Initial state} A flat stove is placed in the right region of the kitchen table, with its knob turned off. A frying pan is placed in the central region of the table, and a white bowl is placed on the tabletop just behind the pan. A wooden two-layer shelf is placed along the left edge of the table.

\begin{figure*}[thbp]
        \centering
    \includegraphics[width=1\textwidth]{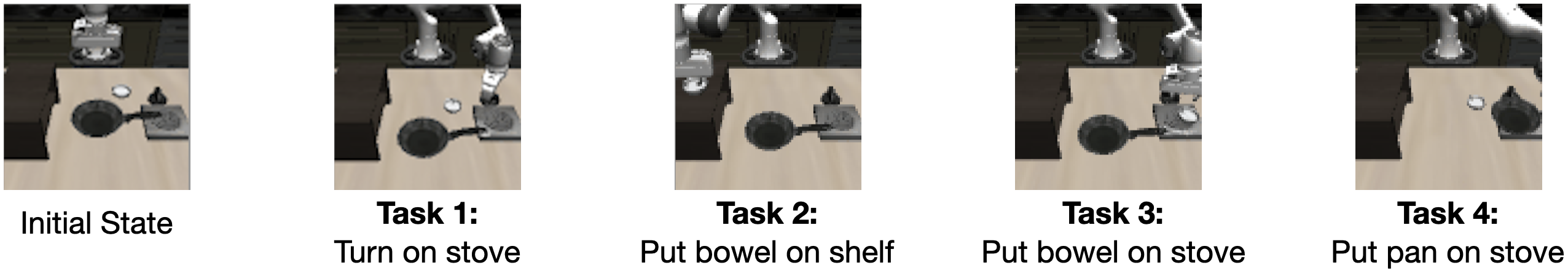}
\caption{Short horizon tasks in LIBERO simulation.}
\label{fig:short_tasks}
\vspace{-1pt}
\end{figure*}

\paragraph{Short Horizon Tasks} There are 4 tasks chosen for short-horizon evaluation: Task-1: turn on the stove; Task-2: put bowel on shelf
      ; Task-3: put bowel on stove
      ; Task-4: put pan on stove. The goal setups are visualized in Figure~\ref{fig:short_tasks}

\begin{figure*}[thbp]
        \centering
    \includegraphics[width=1\textwidth]{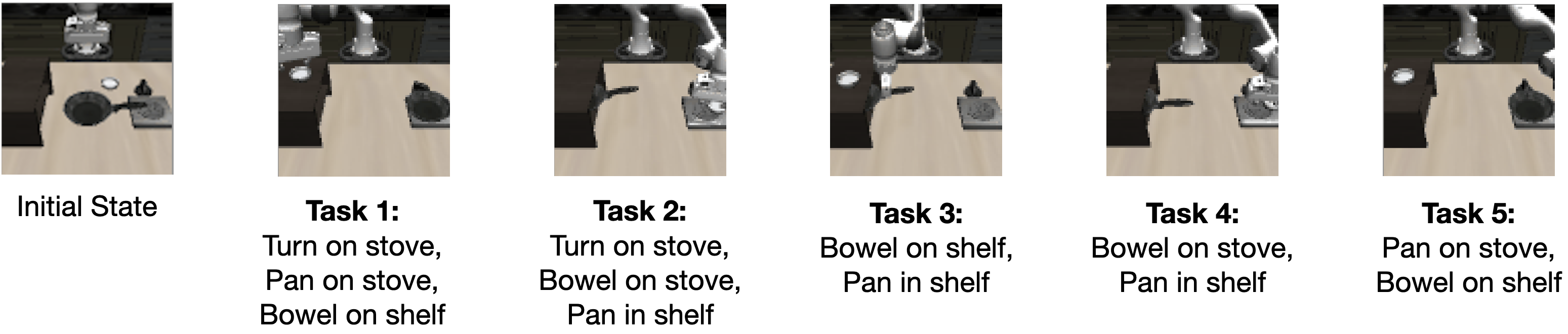}
\caption{Long horizon tasks in LIBERO simulation.}
\label{fig:long_tasks}
\vspace{-1pt}
\end{figure*}

\paragraph{Long Horizon Tasks} There are five tasks chosen for short‐horizon evaluation: Task‐1: turn on stove, pan on stove, bowl on shelf; Task‐2: turn on stove, bowl on stove, pan in shelf; Task‐3: bowl on shelf, pan in shelf; Task‐4: bowl on stove, pan in shelf; Task‐5: pan on stove, bowl on shelf. The goal setups are visualized in Figure~\ref{fig:long_tasks}

\subsection{Real World Setup}

 \begin{figure*}[thbp]
        \centering
    \includegraphics[width=0.8\textwidth]{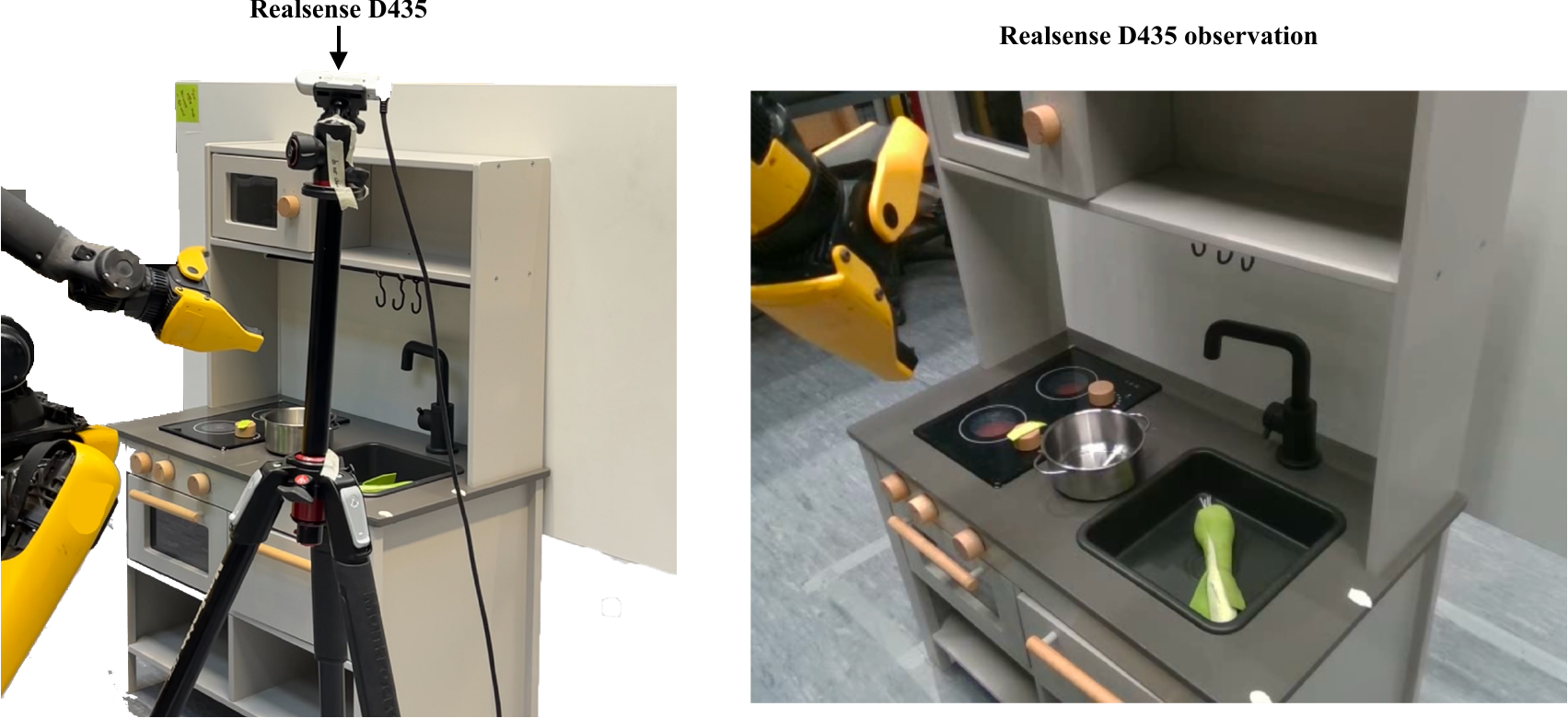}
\caption{Real kitchen setup.}
\label{fig:real_setup}
\vspace{-1pt}
\end{figure*}

\subsubsection{Setups}
Figure~\ref{fig:real_setup} illustrates the real-world task setup. We design kitchen object manipulation tasks for the Boston Dynamics Spot robot arm with a control frequency of 7Hz. \textbf{1) Observation}: We use a Realsense D435 camera to receive a third-person view $512\times512$ RGB observation. 
\textbf{2) Action space}: The action at time step \(t\) is defined as
\[
    a_t = \bigl[\Delta x_t,\;\Delta y_t,\;\Delta z_t,\;\Delta \phi_t,\;\Delta \theta_t,\;\Delta \psi_t,\;g_t\bigr]
    \;\in\;\mathbb{R}^6 \times [0,1],
\]
where
\begin{itemize}
  \item \(\Delta x_t,\Delta y_t,\Delta z_t\) are the Cartesian displacement increments of the end-effector,
  \item \(\Delta \phi_t,\Delta \theta_t,\Delta \psi_t\) are the roll--pitch--yaw rotation increments of the end-effector,
  \item \(g_t\in[0,1]\) is the gripper's opening degree (0 = fully closed, 1 = fully open).
\end{itemize}
All pose increments \(\Delta x,\dots,\Delta \psi\) are expressed in the Spot robot's central coordinate frame. This 6D delta-pose plus gripper-degree parameterization enables precise, relative control of the manipulator in reinforcement learning.

 \begin{figure*}[thbp]
        \centering
    \includegraphics[width=1\textwidth]{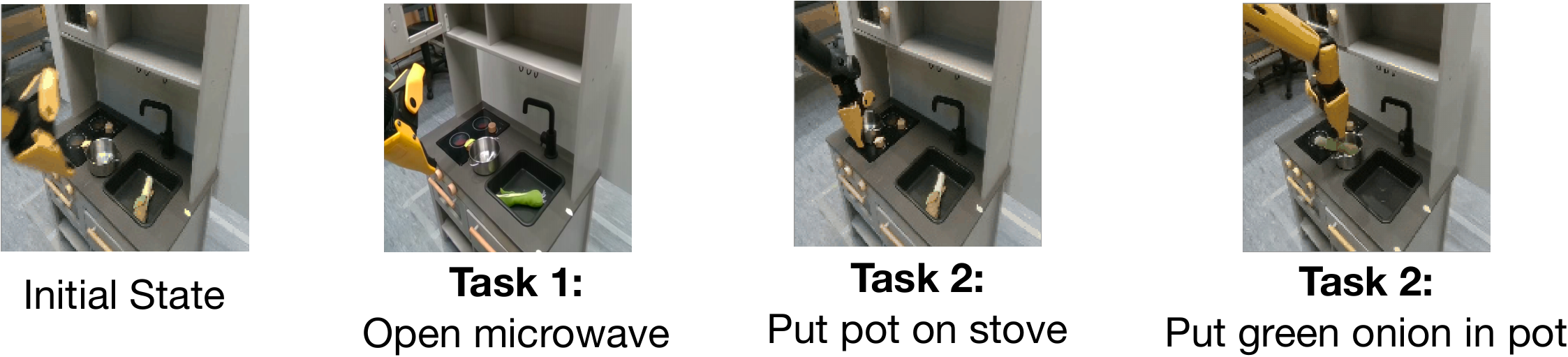}
\caption{Short horizon tasks in real world.}
\label{fig:real_short_tasks}
\vspace{-1pt}
\end{figure*}

 \begin{figure*}[thbp]
        \centering
    \includegraphics[width=1\textwidth]{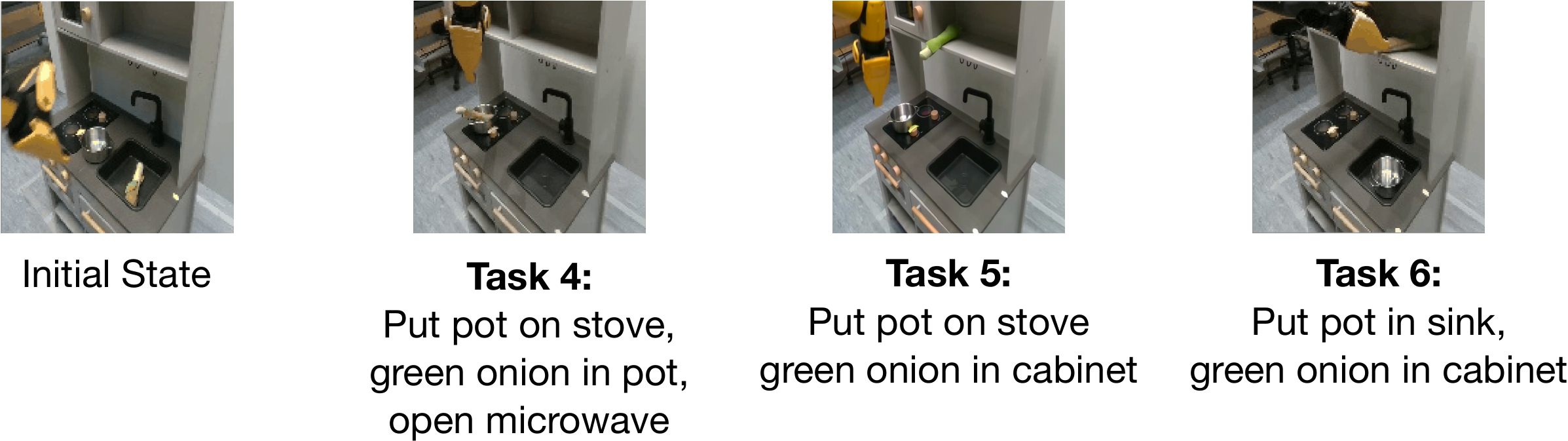}
\caption{Long horizon tasks in real world.}
\label{fig:real_long_tasks}
\vspace{-1pt}
\end{figure*}

\subsubsection{Tasks}
For each task, the initial state and subgoals are pre‐defined.
\paragraph{Initial state} A green onion is put in the sink, microwave on the top-left corner is closed, a pot is placed in between the stove and the sink.

\paragraph{Short Horizon Tasks}
There are three tasks chosen for short‐horizon evaluation in the real setup: Task-1: Put pot on stove; Task-2: Put green onion in pot; Task-3: Open microwave. The goal setups are visualized in Figure~\ref{fig:long_tasks}.

\paragraph{Long Horizon Tasks}
There are three tasks chosen for long‐horizon evaluation in the real setup: Task‐4: Put pot on stove, green onion in pot, open microwave; Task‐5: Put pot on stove, green onion in cabinet; Task‐6: Put pot in sink, green onion in cabinet. The goal setups are visualized in Figure~\ref{fig:real_long_tasks}.  The whole task is completed if and only if all subgoals are completed in the correct order.

\subsection{Demonstration collections}
\textbf{1) LIBERO kitchen demonstrations}: We use the demonstrations collected by Mimicplay~\cite{Mimicplay}. The Mimicplay dataset contains 33 human play demonstrations averaging 7 minutes each (30 frames per second). 
\textbf{2) Real kitchen demonstrations}: The robot teleoperation data is collected using an OptiTrack motion capture system~\cite{optitrack2025}, which tracks the 6D pose of a gripper as it is operated by a human demonstrator. The control frequency of the robot arm and gripper is 7Hz. We collected 15 demonstrations in total: on average, the human operator interacts with the scene for 20 minutes per demonstration (5 frames per second). 
In both simulation and real-world setups, the human operator is asked to explore the environment by randomly teleoperating the robot arm to conduct different tasks based on the operator's own intention.

\section{Vis2Plan details}

\subsection{Details of skill segmentation}

Here we provide the  additional details of segmenting skills from unlabeled demonstration videos for Section~\ref{sec:extract_skills}.

\textbf{Video preprocessing}\quad 
Here we provide detailed explanation for Figure~\ref{fig:extract_skills} (1). Given a demonstration video and a fixed set of \(K\) task‑related object categories \(\{o_1, \dots, o_K\}\), we first detect each \(o_k\) in the initial frame using the open‑vocabulary VLM Qwen~\cite{QWEN} and then track these bounding boxes across time with SAM2~\cite{SAM2}, yielding \(K\) object‑centric videos. Now the demonstration video $V_\text{demo}$ can be described as a combination of object-centric videos $V_k$:
\(
V_\text{demo} = \bigl(V_{1},\dots,\,V_{k}\bigr). \quad V_k = \bigl(I_{k,1},\dots,\,I_{k,T}\bigr),
\)
where $I_{k,i}$ is the object-centric frame and \(T\) is the total number of demonstration frames.
Next,  we extract a feature trajectory from each clip via a vision foundation model \(\mathbf{F}_{\mathrm{VFM}}\) (e.g., SigLIP2~\cite{siglip2}):
\[
\mathbf f^d_k
= \bigl(\mathbf{F}_{\mathrm{VFM}}(I^d_{k,1}),\,\mathbf{F}_{\mathrm{VFM}}(I^d_{k,2}),\,\dots,\,\mathbf{F}_{\mathrm{VFM}}(I^d_{k,T_k})\bigr)
\;\in\;(\mathbb R^D)^{T},
\]
where \(D\) is the feature dimensionality.  In this way, the unlabeled visual demonstration is represented by the set of \(K\) feature sequences:
\(
\mathcal F^d = \{\mathbf f^d_1, \mathbf f^d_2, \dots, \mathbf f^d_K\}
\).

\begin{figure*}[thbp]
    \centering
    \includegraphics[width=1\textwidth]{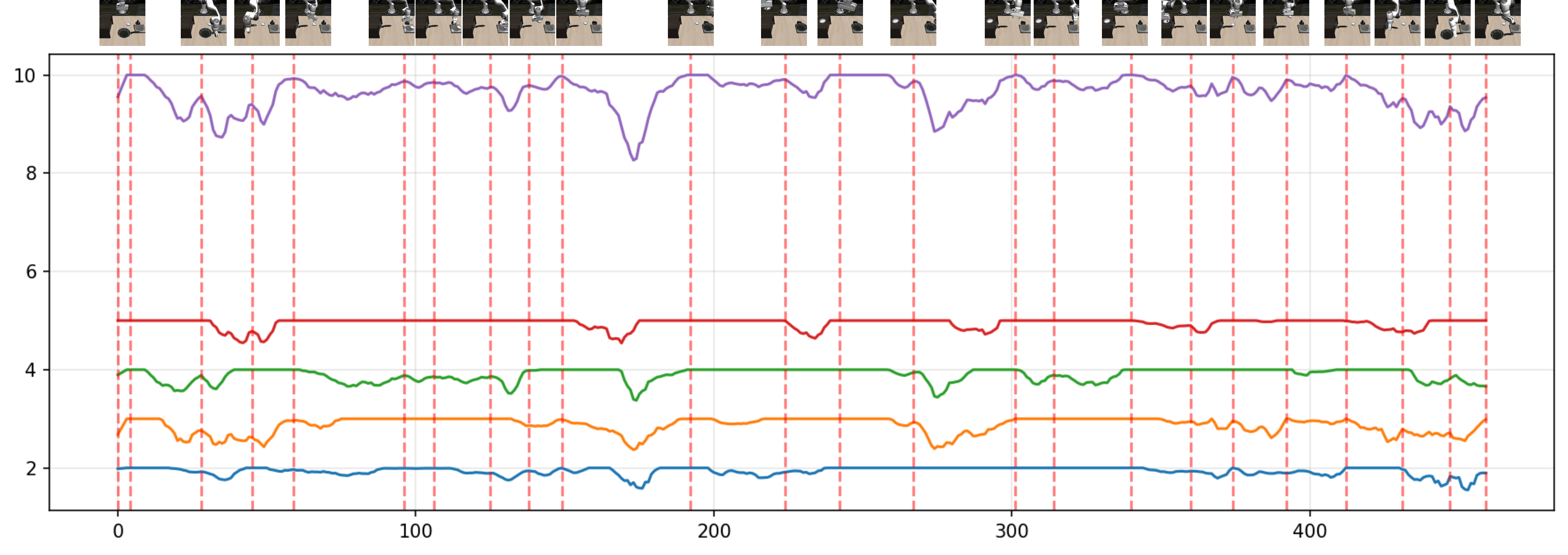}
\caption{\textbf{Skill segmentaion visualization}: Here we provide on skill segmentation (stable state identification) result of one demonstration. The line plots are the visualization of object-centric adjacent similarities and the summation of object-centric adjacent similarities $\mathbb{S}_t$ (the top purple line).  The red vertical dash lines are the NMS peak-finding results of $\mathbb{S}_t$, and the corresponding stable states (image frames of the peaks) are plot on top of the figure.}
\label{fig:skill_segmentation}
\vspace{-15pt}
\end{figure*}

\textbf{Stable State Identification}\quad Here we provide detailed explanation for Figure~\ref{fig:extract_skills} (2):
We define a stable state is where all objects' visual status remain static or similar, and a skill is a transition between two stable states.
Our assumption is that the object's visual features can describe the objects' status (e.g., if it is under manipulation or not), and we can detect if object is under manipulation by tracking  the objects' visual appearance.  Follow this motivation,  we extract segment video using equations~\cite{motion_segment}:
\begin{equation}
    \mathbb{S}_\text{t} =  \sum_{i=1}^{K} \left( \frac{\mathbf{F}_{\mathrm{VFM}}(I^d_{i,t}) \cdot \mathbf{F}_{\mathrm{VFM}}(I^d_{i,t+1})}{\|\mathbf{F}_{\mathrm{VFM}}(I^d_{i,t})\| \|\mathbf{F}_{\mathrm{VFM}}(I^d_{i,t+1})\|}  \right), \quad \mathbb{K}=\mathrm{NMS}(\frac{\sum\limits_{i=t-k}^{t+k} W(\mathbb{S}_t, \mathbb{S}_i)\mathbb{S}_i}{\sum\limits_{i=t-k}^{t+k} W(\mathbb{S}_t, \mathbb{S}_i)})
\end{equation}
Here, $\mathbb{S}_t$ is the temporal similarity at time $t$, summing $t$ and $t+1$ frame's object-centric cosine similarities. After Gaussian smoothing, we apply non-maximum suppression (NMS) within a local temporal window and select the highest peaks as key stable frames $\mathbb{K}$ (see Figure~\ref{fig:extract_skills} (2)). Two neighboring stable state can be used to describe one sub-skill.

\begin{figure*}[thbp]
    \centering
    \includegraphics[width=1\textwidth]{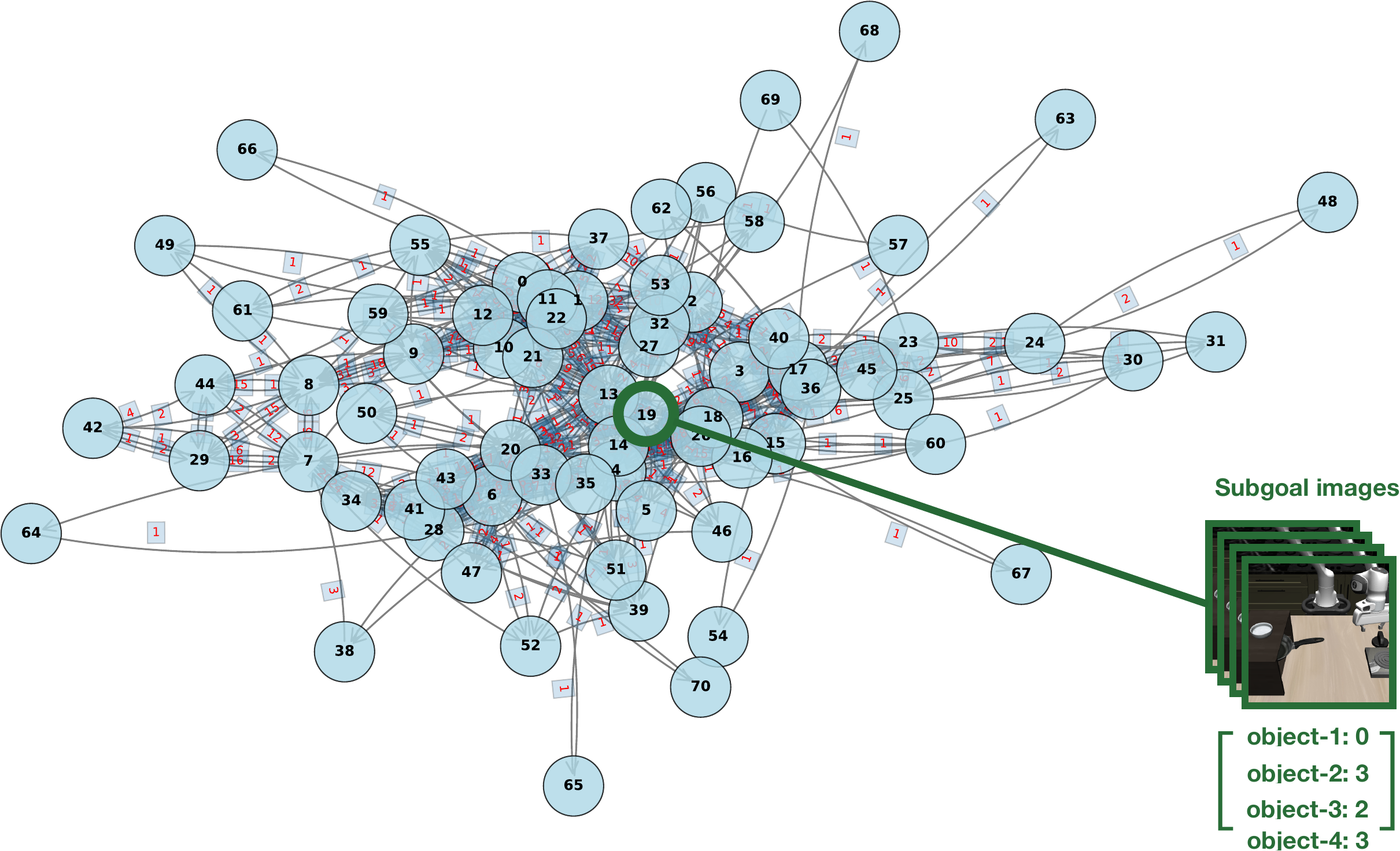}
    \caption{LIBERO kitchen symbolic graph example. Each node of the graph is described by the symbolic representation of task-relevant objects. We also provide one symbolic node example (node 19) to illustrate the node structure.}
    \label{fig:sym_graph}
    \vspace{-1pt}
\end{figure*}

\textbf{Symbolic Graph}\quad 
As described in Section~\ref{sec:extract_skills}, we use SigLIP2 features with unsupervised learning to extract the symbolic representations of key stable states. Using these representations, we construct a symbolic graph for high-level planning. Figure~\ref{fig:sym_graph} presents a visualization of the symbolic graph for the LIBERO kitchen task.

\subsection{Network implementations}
 \begin{figure*}[thbp]
        \centering
    \includegraphics[width=1\textwidth]{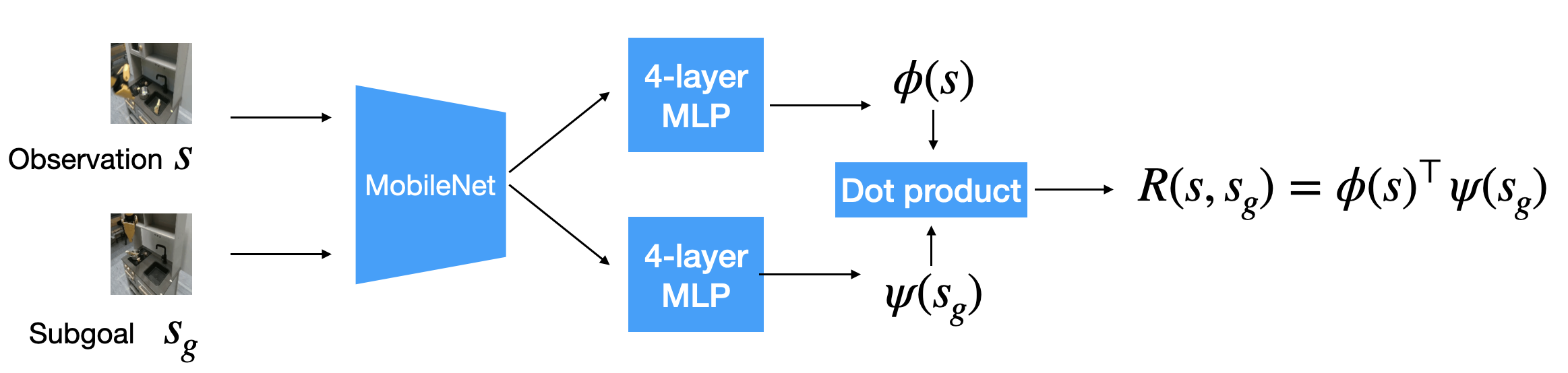}
\caption{Overview of contrastive RL value network architecture (reachability network). We follow the implmentation suggestions in work ~\cite{CRL2} and modify it to state-only variant. }
\label{fig:reachability_arch}
\vspace{-1pt}
\end{figure*}
\paragraph{Reachability Estimator $R(s,s_g)$} We follow the methodology from contrastive RL work~\cite{CRL2} to implement our reachability estimator. We modify the Contrastive RL approach to distinguish between a future state sampled from the average discounted state-occupancy measure in the state-only case:
\[
  s^+_f \sim p^\pi(s_{t+}\mid s)
    = \int\int p^{\pi(\cdot\mid g)}(s_{t+}\mid s,a)\,p^\pi(g\mid s,a)\,\pi(a|s)\,\mathrm{d}g \mathrm{d}a,
\]
and a future state sampled from an arbitrary state–action pair:
\[
  s^-_f \sim p(s_{t+})
    = \int p^\pi(s_{t+}\mid s)\,p(s)\,\mathrm{d}s\,
\]

We train the estimator using the NCE-Binary objective:
\begin{equation}\label{eq:nce_binary}
\begin{aligned}
\mathcal{L}
=&\;\mathbb{E}_{s^+_f\sim p^\pi(s_{t+}\mid s)}\Bigl[\log\sigma\bigl(\phi(s)^\top\psi(s^+_f)\bigr)\Bigr]
+\mathbb{E}_{s^-_f\sim p(s_{t+})}\Bigl[\log\bigl(1-\sigma\bigl(\phi(s)^\top\psi(s^-_f)\bigr)\bigr)\Bigr].
\end{aligned}
\end{equation}

After training, the inner product $R(s, s_g)=\phi(s)^\top\psi(s_g)$ between the state representation $\phi(s)$ and future (goal) state $\psi(s_g)$ is proportional to the state-occupancy measure and can be used as a reachability estimator.
Figure~\ref{fig:reachability_arch} illustrates the reachability network architecture. For training hyperparameters, we strictly follow those used in \cite{CRL2}.

\begin{table*}[t]
  \centering
  \scriptsize
  \begin{minipage}[t]{0.32\textwidth}
    \centering
    \captionof{table}{Low‐level Policy Hyperparameters: AdaFlow.}
    \label{tab:adaflow_hyperparams}
    \begin{tabular}{@{}ll@{}}
      \toprule
      \textbf{Hyperparameter}             & \textbf{AdaFlow}      \\ 
      \midrule
      Learning rate                       & $1\times 10^{-4}$           \\
      Optimizer                           & AdamW                       \\
      $\beta_1$                           & 0.95                        \\
      $\beta_2$                           & 0.999                       \\
      Weight decay                        & $1\times 10^{-6}$           \\
      Batch size                          & 64                          \\
      Epochs                              & 500     \\
      LR scheduler                        & cosine                      \\
      EMA decay rate                      & 0.9999                      \\
      \addlinespace
      \(\eta\)                            & 1.0                         \\
      \(\epsilon_{\min}\)                & 10                          \\
      \addlinespace
      Action prediction horizon           & 16                          \\
      Observation inputs               & 2                           \\
      Action execution horizon            & 8                           \\
      Obs.\ input size                    & \(128\times128\)            \\
      \bottomrule
    \end{tabular}
  \end{minipage}\hfill
  \begin{minipage}[t]{0.32\textwidth}
    \centering
    \captionof{table}{Low‐level Policy Hyperparameters: Mimicplay (latent).}
    \label{tab:hp_latent_planner}
    \begin{tabular}{@{}ll@{}}
      \toprule
      \textbf{Hyperparameter}            & \textbf{Default}             \\ 
      \midrule
      Batch Size                         & 16                           \\
      Learning Rate (LR)                 & \(1\times10^{-4}\)           \\
      Num Epoch                          & 1000                         \\
      LR Decay                           & None                         \\
      KL Weights \(\lambda\)             & 1000                         \\
      MLP Dims                           & [400, 400]                   \\
      Img.\ Encoder – Left View          & ResNet-18                    \\
      Img.\ Encoder – Right View         & ResNet-18                    \\
      Image Feature Dim                  & 64                           \\
      GMM Num Modes                      & 5                            \\
      GMM Min Std                        & 0.0001                       \\
      GMM Std Activation                 & Softplus                     \\
      \bottomrule
    \end{tabular}
  \end{minipage}\hfill
  \begin{minipage}[t]{0.32\textwidth}
    \centering
    \captionof{table}{Low‐level Policy Hyperparameters: Mimicplay (action).}
    \label{tab:hp_robot_policy}
    \begin{tabular}{@{}ll@{}}
      \toprule
      \textbf{Hyperparameter}            & \textbf{Default}             \\ 
      \midrule
      Batch Size                         & 16                           \\
      Learning Rate (LR)                 & \(1\times10^{-4}\)           \\
      Num Epoch                          & 1000                         \\
      Train Seq Length                   & 10                           \\
      LR Decay Factor                    & 0.1                          \\
      LR Decay Epoch                     & [300, 600]                   \\
      MLP Dims                           & [400, 400]                   \\
      Img.\ Encoder – Wrist View         & ResNet-18                    \\
      Image Feature Dim                  & 64                           \\
      GMM Num Modes                      & 5                            \\
      GMM Min Std                        & 0.01                         \\
      GMM Std Activation                 & Softplus                     \\
      GPT Block Size                     & 10                           \\
      GPT Num Head                       & 4                            \\
      GPT Num Layer                      & 4                            \\
      GPT Embed Size                     & 656                          \\
      GPT Dropout Rate                   & 0.1                          \\
      GPT MLP Dims                       & [656, 128]                   \\
      \bottomrule
    \end{tabular}
  \end{minipage}
\end{table*}

\textbf{Low-level goal-conditioned policies}\quad The low-level policy should generate a sequence of actions to achieve the subgoal proposed by the visual planner. It receives the current image observation and a goal image to generate appropriate actions. All goal-conditioned policies are trained with a hindsight relabeling strategy: given an unlabeled video demonstration, the goal image \(g^r_t\in \mathcal{V}^r\) is defined as the frame occurring \(H\) steps after the current time step in the demonstration. Here, \(H\) is drawn uniformly from the integer range \([100,600]\) (i.e., 5–30 seconds), serving as a data-augmentation mechanism. We adapt AdaFlow~\cite{Adaflow} for LIBERO simulation and Mimicplay~\cite{Mimicplay} for real-world tasks as our low-level goal-conditioned policies.
\begin{itemize}
    \item \textbf{Goal-conditioned AdaFlow}\quad In LIBERO simulation, we implement a goal-conditioned variant of the imitation-flow-based generative policy from AdaFlow~\cite{Adaflow} as our low-level controller. At each step, this policy receives the current observation and the target subgoal image (both at $128\times128$ resolution) and predicts a horizon of 16 consecutive actions. All architecture and hyperparameters remain the same as implemented in ~\cite{Adaflow}. In Table \ref{tab:adaflow_hyperparams}, we summarize all key settings of the goal-conditioned AdaFlow policy.
    
    \item \textbf{Mimicplay Policy}\quad In real-world tasks, we use the Mimicplay policy as the low-level control policy. We maintain the implemented architecture and hyperparameters as used in \cite{Mimicplay}. At each time step $t$, the Mimicplay policy receives the current image observation and the next subgoal image observation (both at $128\times128$ resolution) and outputs an 16-step end-effector pose trajectory, which conditions the low-level policy to generate a sequence of 16-step actions for execution. Table \ref{tab:hp_latent_planner} summarizes the hyperparameters used to train the Mimicplay latent planner, and Table \ref{tab:hp_robot_policy} those for the robot policy $\pi$. Parameters prefixed with "GMM" refer to the MLP-based Gaussian mixture model, whereas those prefixed with "GPT" specify the transformer's configuration\cite{Mimicplay}.
\end{itemize}

\subsection{Pseudocode of Closed-loop Symbolic-guided Visual Planning}

\begin{figure*}[thbp]
    \centering
    \includegraphics[width=1\textwidth]{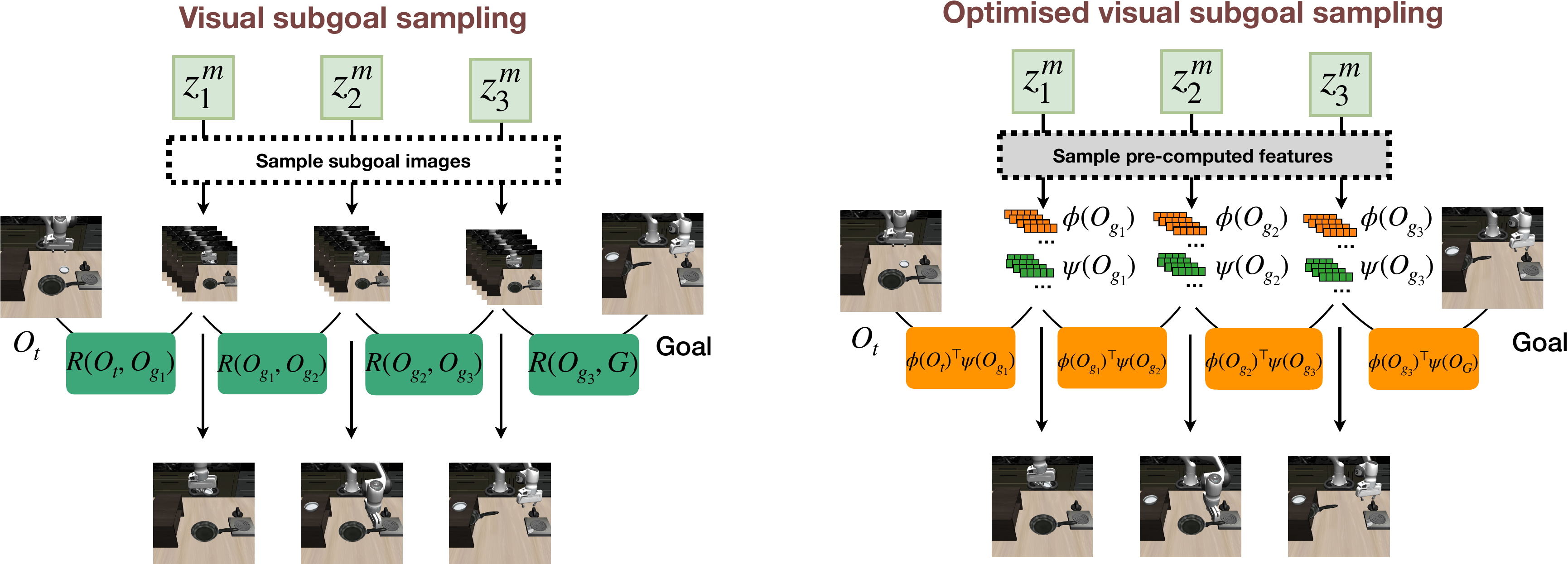}
    \caption{Optimized visual subgoal sampling (stage 2 in Algorithm 1). The right plot illustrates the standard visual subgoal sampling pipeline, while the left plot shows our optimized variant. We directly sample pre-computed reachability features of subgoal images to search for the most feasible visual plan, significantly reducing computational overhead.}
    \label{fig:optimized_vis_sample}
    \vspace{-1pt}
\end{figure*}

Here we provide pseudocode to illustrate how symbolic-guided visual planning is executed, as additional details for Section~\ref{sec:stage3}. More specifically, we optimized the visual subgoal generation step to accelerate visual plan generation. We use the pretrained reachability estimator to calculate the reachability features of all key image frames and save these features with their corresponding images. During visual subgoal sampling, instead of sampling image batches for each symbolic subgoal, we directly sample the subgoal images' reachability feature batch ($\phi(s)$ and $\psi(s)$) to avoid computationally expensive reachability estimation. This optimized sampling approach is visualized in Figure~\ref{fig:optimized_vis_sample}.

\begin{algorithm}[H]
\caption{Stage 3: Symbolic-guided Visual Planning with Vis2Plan}
\label{alg:vis2plan}
\textbf{Input}: Task goal $z_g$, state predictor $C_\theta$, reachability estimator $R_\psi$, goal-conditioned policy $\pi_g$ \\
\textbf{Initialize}: symbolic path $\tau_z = \emptyset$, visual plan $\tau_O = \emptyset$ \\
\textbf{while} goal $z_g$ is not reached \textbf{do} \\
\quad Receive current observation $O_t$ \\
\quad \textbf{// Stage 1: Symbolic Planning} \\
\quad Predict next symbolic state candidates: $\mathcal{Z}_\text{next} = C_\theta(O_t)$ \\
\quad for $z_1$ in $\mathcal{Z}_\text{next}$ do \\
\quad \quad Find symbolic path: $\text{path} = \textsc{Search}(z_1, z_g)$ \\
\quad \quad Update if shorter: $\tau_z = \text{path}$ if $\text{cost}(\text{path}) < \text{cost}(\tau_z)$ \\
\quad \\
\quad \textbf{// Stage 2: Visual Subgoal Generation} \\
\quad for $z_i$ in $\tau_z$ do \\
\quad \quad Sample visual observation: $O_i \sim P_D(\cdot \mid z_i)$ \\
\quad \quad Add to visual plan: $\tau_O = \tau_O \cup \{O_i\}$ \\
\quad Optimize visual plan: $\tau_O = \arg\max_{\tau'_O} \sum_{i=0}^{H-1} R_\psi(O'_i, O'_{i+1})$ using beam search \\
\quad \quad \quad \quad \quad \quad \text{s.t.} $R_\psi(O'_i, O'_{i+1}) \leq \delta, \forall i$ \\
\quad \\
\quad \textbf{// Stage 3: Goal-conditioned Action Generation} \\
\quad Get next visual subgoal: $O_g = \tau_O[1]$ \\
\quad Generate action chunk: $u = \pi_g(O_t, O_g)$ \\
\quad Execute action chunk $u$ in environment \\
\textbf{Return}: task completed
\end{algorithm}

\subsection{End2End Goal-Conditioned Baselines}

\textbf{Goal-conditioned transformers (GC-Transformer)}\quad We use the goal-conditioned transformer architecture defined in Mimicplay as the baseline implementation~\cite{Mimicplay}. The Mimicplay GC-transformer embed the wrist token $w_t$, environment embedding $e_t$, and proprioceptive embedding $p_t$ to form the sequence
\(
  s_{t:t+T} = [w_t, e_t, p_t, \dots, w_{t+T}, e_{t+T}, p_{t+T}],
\)
with $T=10$. This sequence is processed by a GPT-style transformer $f_{\mathrm{trans}}$ comprising $N=4$ layers of multi-head self-attention (4 heads) and position-wise feed-forward networks with intermediate dimension 656 and output dimension 128. The model uses a block size of 10, embedding dimension of 656, and dropout rate of 0.1. Given the past $T-1$ embeddings, $f_{\mathrm{trans}}$ autoregressively predicts the next action feature $x_T$. The final control command $a_t$ is produced by passing $x_t$ through a two-layer MLP with hidden dimensions [400,400]. Visual observations from wrist, left, and right camera views are encoded via ResNet-18 into 64-dimensional features. To model the multimodal action distribution, we employ an MLP-based Gaussian mixture model with 5 components and a minimum standard deviation of 0.01 (Softplus activation). We train the network for 1000 epochs with a batch size of 16 and initial learning rate $1\times10^{-4}$, decayed by a factor of 0.1 at epochs 300 and 600.

\textbf{Goal-conditioned Diffusion Policy (GC-Diffusion)} We adapt the Goal-conditioned variant diffusion policy from work Adaflow \cite{Adaflow} as one baseline.  
The diffusion‐based robot policy is trained with the AdamW optimizer (learning rate $1\times10^{-4}$, $\beta_1=0.95$, $\beta_2=0.999$, weight decay $1\times10^{-6}$), using a batch size of 64. We train for 500 epochs in the likelihood‐weighting phase and 3000 epochs in the reverse‐mapping phase, employing a cosine learning‐rate scheduler and an exponential moving‐average decay of 0.9999. The denoising diffusion model uses 100 forward diffusion steps and 100 inference steps under the DDPM framework. At test time, the policy predicts an action horizon of 16 steps conditioned on the two most recent observations, executes 8 actions per cycle, and processes visual inputs at a resolution of $128\times128$.

\paragraph{Goal-conditioned Action-Chunk Transformer (GC-ACT)}\quad 
Action-chunk transformer (ACT) was proposed in \cite{aloha}. For our work, we implemented a goal-conditioned variant of ACT based on the code repository available at \url{https://github.com/Shaka-Labs/ACT}. Our GC-ACT receives the current image observation and a goal image observation as inputs, and generates a 16-step action chunk for execution.

\subsection{High-level Planner Baselines}

\paragraph{AVDC} We adapt the video diffusion generation model from \cite{AVDC} as one of our visual planner baselines. We strictly follow the configuration provided in the AVDC public code repository (\url{https://github.com/flow-diffusion/AVDC}). Specifically, we modify the original implementation to use an image goal instead of a task text description. The AVDC diffusion video generation model receives the current image observation and a goal image observation (both at $128\times128$ resolution) and outputs a 9-step horizon image plan, with each subgoal image at $64\times64$ resolution. 
We train the AVDC video generation model similarly to training a goal-conditioned policy: we hindsight sample the future image of the current trajectory as the goal image, and subsample the intermediate frames to 9 frames as the subgoals (target video frames). Given an unlabeled video demonstration, the goal image \(g^r_t\in \mathcal{V}^r\) is defined as the frame occurring \(H\) steps after the current time step in the demonstration. Here, \(H\) is drawn uniformly from the integer range \([100,1500]\) (i.e., 5–75 seconds).

\paragraph{GSR} We implemented the baseline GSR follow the instruction from ~\cite{gsr}, where the author build a directed graph from offline demonstrations. Based on GSR implementation, each image observation from demonstrations is treated as a vertex of a graph, and the directed edge between vertexes indicate the image state transition. There are two types of edges built in GSR: 1) Dataset edge: represents ground truth transitions on each demonstration trajectory $(v_0, v_1), (v_1, v_2), ...$, serving as an approximation of the world dynamics; 2) Augmented edge: a bidirectional edge between nodes $u$ and $v$ is added if they both lie in the tolerance range of each other in the pretrained representation space, bridging similar states across trajectories.  In this work, we use our trained reachability estimator $R(s,s_g)$ as the pretrained representation network to construct the augmented edges.

\paragraph{UVD-graph} Here, we develop a graph-based approach using the UVD implementation. We finetune the UVD feature extraction following the instructions in \cite{UVD} (\url{https://github.com/zcczhang/UVD/}). We then use UVD to segment the demonstration into sub-tasks. Similar to GSR, we use our pretrained reachability estimator $R(s,s_g)$ to build a graph based on the segmented subgoals. This implementation creates a more sparse graph compared to GSR. The key difference between UVD-graph and our Vis2Plan is that Vis2Plan uses object-centric features to identify subgoals and creates a symbolic description of key states, enabling more structured and interpretable planning.

\section{Visual Planning Results}

\begin{table}[h]
  \centering
  \caption{Average visual plan generation speed (seconds) for different planners across task types. "Sim" refers to LIBERO simulation and "Real" refers to real-world tasks. AVDC shows consistent visual plan generation times across tasks because it generates a fixed number of subgoals (9) regardless of task complexity.}
  \label{tab:plan_speed}
  \begin{tabular}{lcccc}
    \toprule
    \textbf{Planner}     & \textbf{Sim Short} & \textbf{Sim Long} & \textbf{Real Short} & \textbf{Real Long} \\
    \midrule
    GSR                  & 0.10               & 15.12             & 0.11                & 10.13              \\
    UVD-graph            & 1.2$\times$10$^{-4}$ & 2.8$\times$10$^{-4}$ & 1.1$\times$10$^{-4}$ & 1.8$\times$10$^{-4}$ \\    
    AVDC                 & 1.42               & 1.42              & 1.42                & 1.42               \\
    Vis2Plan (ours)      & 0.03               & 0.04              & 0.03                & 0.05               \\
    \bottomrule
  \end{tabular}
\end{table}

\subsubsection{Visual Planning Inference Speed Comparison}

Here we compare the visual plan generation inference speed across different methods (Table~\ref{tab:plan_speed}). GSR builds a graph based on one-step temporal transitions, where each vertex represents an image frame from the demonstration. This creates an extremely complicated graph that is time-consuming to search, resulting in visual plan generation times of several seconds for long-horizon tasks. AVDC uses a diffusion generative model to generate image sequences, but the diffusion process is computationally expensive, making real-time planning infeasible.

Despite having similar graph complexity to GSR, the UVD-graph approach is the fastest planner as it does not require a visual goal sampling procedure. However, our Vis2Plan framework still achieves near real-time planning performance while maintaining high plan quality and physical feasibility.

\subsection{Visual Plan results}
Here we provide some representative planning results of Vis2Plan in both simulation and real-world tasks. For additional visualization results, please refer to the second supplementary material.

\begin{figure*}[thbp]
    \centering
    \begin{subfigure}{\textwidth}
        \centering
        \includegraphics[width=\textwidth]{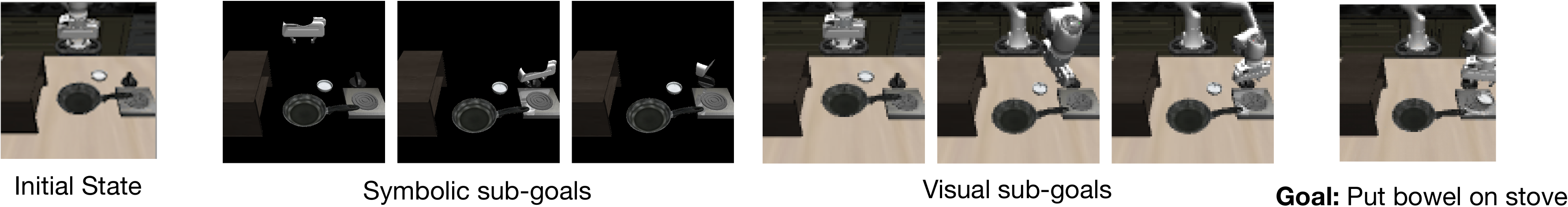}
        \caption{Simulation short-horizon Vis2Plan visual planning example: put bowl on stove}
    \end{subfigure}
    
    \vspace{0.5cm}
    
    \begin{subfigure}{\textwidth}
        \centering
        \includegraphics[width=\textwidth]{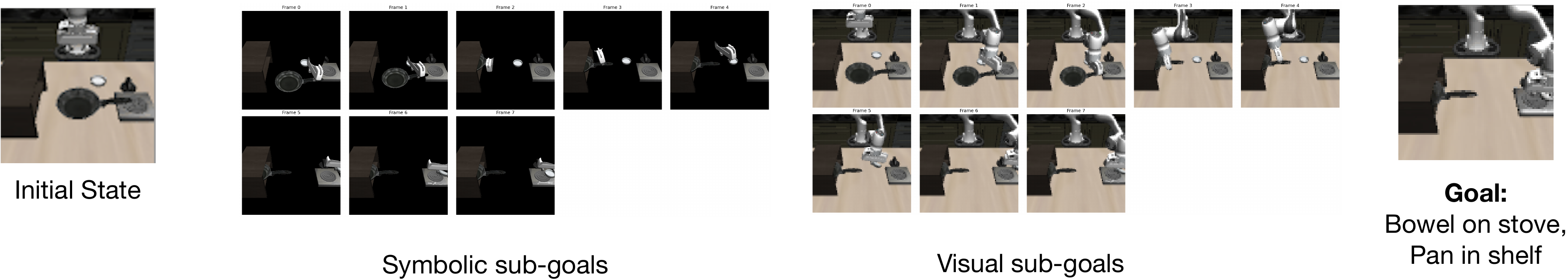}
        \caption{Simulation long-horizon Vis2Plan visual planning example: put bowl on stove and pan in shelf}
    \end{subfigure}
    
    \vspace{0.5cm}
    
    \begin{subfigure}{\textwidth}
        \centering
        \includegraphics[width=\textwidth]{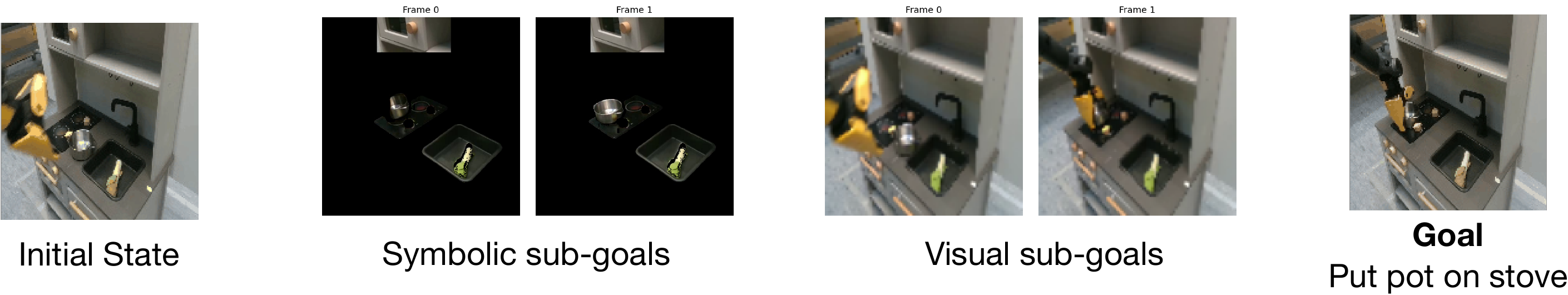}
        \caption{Real world short-horizon Vis2Plan visual planning example: put pot on stove}
    \end{subfigure}
    
    \vspace{0.5cm}
    
    \begin{subfigure}{\textwidth}
        \centering
        \includegraphics[width=\textwidth]{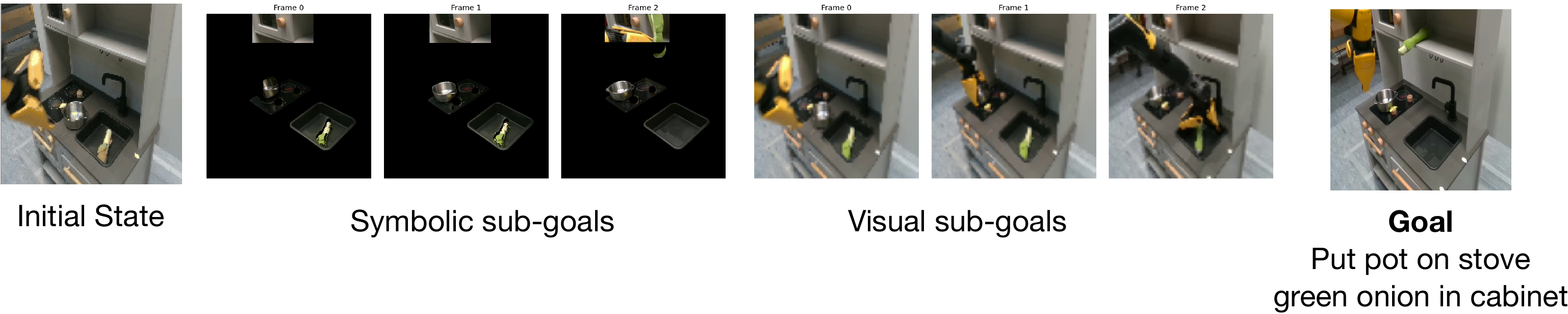}
        \caption{Real world long-horizon Vis2Plan visual planning example: put pot on stove and green onion in cabinet}
    \end{subfigure}
    
    \caption{Representative Vis2Plan visual planning examples in both simulation and real-world environments. (a)-(b) Simulation examples showing short and long-horizon tasks. (c)-(d) Real-world examples demonstrating similar capabilities in physical environments.}
    \label{fig:planning_examples}
    \vspace{-1pt}
\end{figure*}







The choice metric is simple: the feature extracted by the VFM shall have good ability to distinguish  in an unsupervised manner
We design the experiment in this way:


\end{document}